\documentclass{article} 
\usepackage{iclr2027_conference,times}

\usepackage{amsmath,amsfonts,bm}

\def\eqref#1{equation~\ref{#1}}

\def\1{\bm{1}}

\DeclareMathAlphabet{\mathsfit}{\encodingdefault}{\sfdefault}{m}{sl}
\SetMathAlphabet{\mathsfit}{bold}{\encodingdefault}{\sfdefault}{bx}{n}

\usepackage{hyperref}
\usepackage{url}
\usepackage{booktabs}
\usepackage{multirow}
\usepackage{adjustbox}
\usepackage{array}
\usepackage{longtable}
\usepackage{graphicx}
\usepackage{amsmath}
\usepackage{amssymb}
\usepackage{placeins}
\usepackage{needspace}
\usepackage{float}
\usepackage{wrapfig}
\usepackage{xcolor}
\usepackage[most]{tcolorbox}
\usepackage{fvextra}

\newtcolorbox{promptbox}[1][]{
    breakable,
    enhanced,
    colback=gray!4,
    colframe=black!55,
    boxrule=0.4pt,
    arc=1pt,
    left=4pt,
    right=4pt,
    top=4pt,
    bottom=4pt,
    title={#1},
    fonttitle=\bfseries
}

\newcommand{\infobox}[1]{
    \begin{tcolorbox}[
        colback=white!90!gray,
        colframe=teal!60!black,
        arc=5pt, boxsep=5pt,
        left=5pt, right=10pt, top=2pt, bottom=3pt,
        boxrule=0.8pt,
        drop shadow=gray!50!white,
        enhanced jigsaw
    ]
         \textit{#1}
    \end{tcolorbox}
}

\title{\ours{}: Evaluating Continual Learning for Self-Evolving Embodied Agents}

\author{%
\bf
Jie Huang$^{\star,1,2}$ \quad
Yanan Chen$^{\star,1}$ \quad
Ruixun Liu$^{\star,1}$ \quad
Kaichen He$^{1}$ \\
\bf
Xinyi He$^{1}$ \quad
Rui Huang$^{3}$ \quad
Zilong Zheng$^{2}$ \quad
Zhenliang Zhang$^{2}$ \quad
Yiwu Zhong$^{1,2,\dagger}$ \\
\rm
$^{\star}$ Equal contribution \qquad $^{\dagger}$ Corresponding author \\
$^{1}$ School of Intelligence Science and Technology, Peking University \\
$^{2}$ State Key Laboratory of General Artificial Intelligence, BIGAI, Beijing, China \\
$^{3}$ University of Hong Kong
}

\newcommand{\ours}{\textit{EmbodiedCLBench}}

\newif\ifarxivversion
\arxivversiontrue
\ifarxivversion
\iclrfinalcopy
\fi
\begin{document}

\maketitle
\ifarxivversion
\lhead{}
\fi

\begin{abstract}
Continual learning requires models to accumulate past experience and transfer it to new tasks. 
Despite substantial research on continual learning, existing work remains largely confined to non-embodied settings, without extending to embodied environments. 
To address this gap, we introduce \ours{}, the first benchmark in embodied simulation for evaluating the continual learning capability of self-evolving agents. 
With compositional generalization as the core evaluation principle, our benchmark evaluates whether agents can first learn from basic tasks and then recombine the acquired skills to solve complex, unseen advanced tasks. 
Based on \ours{}, we conduct extensive experiments across multiple agent harnesses and learning methods, revealing consistent limitations and distinctive behaviors.
Our results show that while advanced tasks can be hardly solved under zero-shot setting, the experience from basic tasks enables consistent improvement. 
Such benefit from the experience remains stable even if the task complexity increases. However, this improvement does not scale accordingly with the number of learning samples, and current agents cannot extract meaningful information from additional experience.
Collectively, our findings highlight continual learning as a fundamental yet underexplored capability, and our benchmark offers valuable resources for designing effective self-evolving agents in embodied environments.
Our code and dataset are available at \url{https://pku-value-lab.github.io/EmbodiedCLBench-homepage}.
\end{abstract}

\section{Introduction}
\label{sec:intro}
Self-evolving agents~\citep{gaoSurveySelfEvolvingAgents2025,fangComprehensiveSurveySelfEvolving2025} can accumulate experience from past interactions and progressively enhance their task performance without additional human supervision. Such agents are capable of retaining and transferring skills across tasks over time, a capability broadly termed \emph{continual learning}~\citep{zhengLifelongLearningLarge2025}. This capability is particularly important for embodied intelligence~\citep{Liu_2025,yifanEmbodiedAISurvey2025,liuAligningCyberSpace2025}, where agents are expected to operate in complex, ever-changing physical environments that cannot be fully anticipated at training time. 
It is therefore important to develop a principled approach for evaluating the capability to learn from prior interactions and transfers the acquired skills to new tasks.

Existing benchmarks in continual learning and embodied intelligence partially address this evaluation goal. 
Research on continual learning has largely focused on non-embodied settings, including knowledge-intensive tasks~\citep{gaoEvoAgentBenchBenchmarkingAgent2026}, interactive digital environments~\citep{jiangSEAEvalBenchmarkEvaluating2026}, and long-term simulated worlds~\citep{caiBuildingSelfEvolvingAgents2025}. These platforms do not capture the spatial complexity and physical constraints inherent in embodied tasks. 
Meanwhile, embodied benchmarks evaluate capabilities such as scene perception~\citep{wangEmbodiedScanHolisticMultiModal2024}, visual navigation~\citep{andersonVisionandLanguageNavigationInterpreting2018}, robotic manipulation~\citep{guManiSkill2UnifiedBenchmark2023a}, and long-horizon task planning~\citep{shridharALFREDBenchmarkInterpreting2020}, but are not specifically designed to assess continual learning. 
Therefore, there remains a lack of a dedicated benchmark for evaluating whether self-evolving agents can accumulate experience and transfer it to solve novel tasks in embodied environments.

\begin{figure}[t]
\centering
\includegraphics[width=\linewidth]{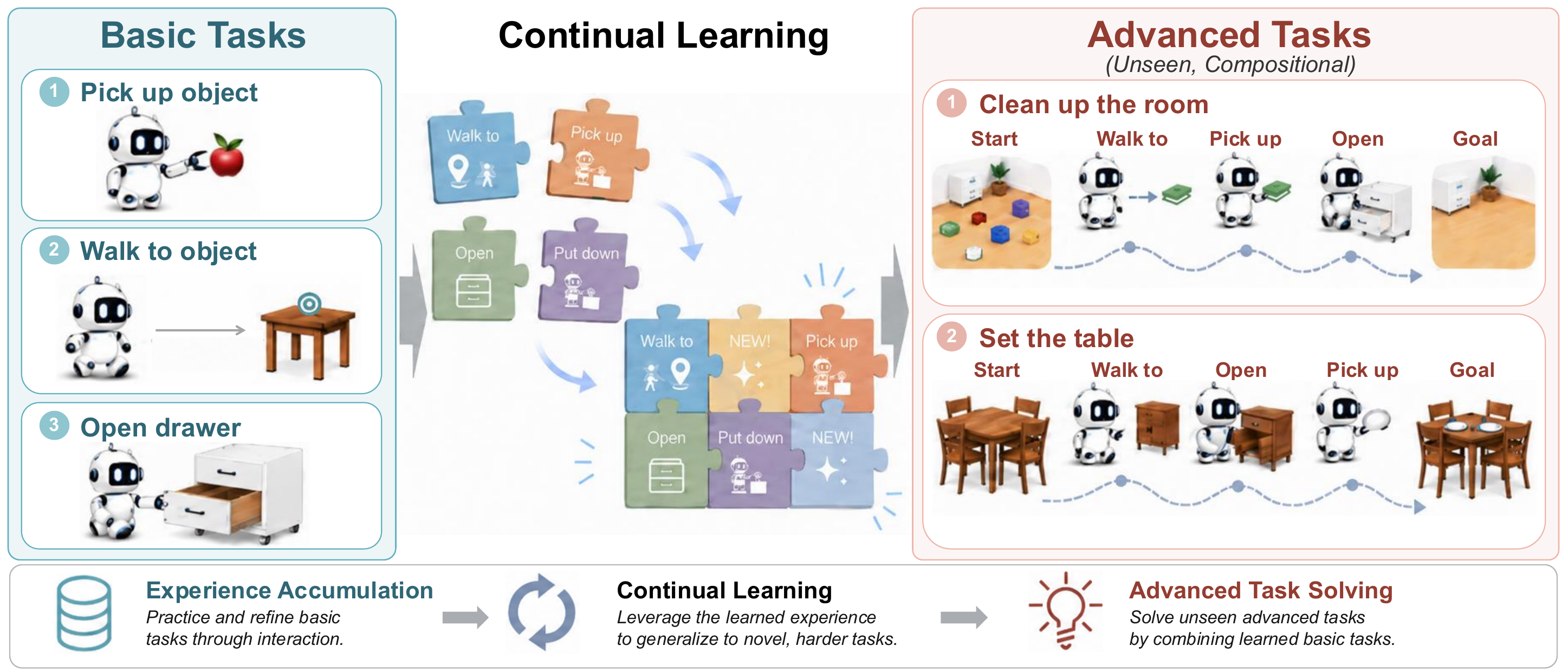}
\vspace{-6mm}
\caption{\textbf{Compositional generalization is leveraged as a principled way to measure continual learning capability.} An agent that has experience from basic tasks should recombine previously acquired skills to solve novel, unseen advanced tasks, rather than learning each new task from scratch.}
\label{fig:intro}
\end{figure}

To address this gap, we introduce \ours{}, the first benchmark in embodied simulation for evaluating the continual learning capability of self-evolving agents.
As illustrated in Figure~\ref{fig:intro}, our design stems from compositional generalization~\citep{furutaExposingLimitationsLanguage2023,guoAtomictoCompositionalGeneralizationMobile2025}. Instead of learning every new task from scratch, a self-evolving agent should recombine previously acquired skill elements to solve novel, unseen tasks.
We thus consider compositional generalization as a principled way to measure agent continual learning capability.
Specifically, we first assign a set of basic tasks for the agent to execute and learn the skills. We then test whether the agent can effectively leverage such experience to solve unseen advanced tasks that require recombining the learned skills.
From this perspective, basic tasks serve as test-time learning~\citep{sun2020test,liu2021ttt++} data for the agent to build experience before attempting advanced tasks.
To this end, \ours{} organizes tasks into three compositional levels: (1)~atomic actions from TongSim~\citep{sunTongSIMGeneralPlatform2025a}, representing basic object--action interactions; (2)~basic tasks, with each composing a small number of atomic actions into practical household operations across multiple categories; and (3)~advanced tasks, with each combining elements from multiple basic-task categories.

Based on \ours{}, we conduct extensive evaluation across multiple agent harnesses (e.g., \textsc{OpenClaw} and \textsc{Codex}) and learning methods (In-Context Learning and Skill Learning). Our results reveal several notable and, in some cases, unexpected findings.
\textit{First}, while advanced tasks can hardly be solved in the zero-shot setting, agents with experience from related basic tasks achieve substantially higher success rates. This validates that experience from simpler tasks can be effectively transferred to unseen, harder tasks.
\textit{Second}, the performance gain over zero-shot agents remains at a stable level as task complexity increases. This suggests that continual learning capability is not sensitive to task difficulty.
\textit{Third}, the improvement from learning does not scale accordingly with the number of learning samples. A single basic task significantly improves performance, and additional samples yield no consistent further improvement.
\textit{Fourth}, past experience facilitates the decision behavior of agents by selecting more meaningful actions with fewer unexecutable ones.

Our contributions are summarized as follows:
\begin{itemize}
    \item We introduce \ours{}, \textbf{the first benchmark} in embodied simulation for evaluating the continual learning capability of self-evolving agents.
    \item \textbf{Our key innovation} lies in leveraging compositional generalization as a principled way to construct tasks, design evaluation protocols, and evaluate continual learning capability.
    \item We conduct a \textbf{comprehensive evaluation} across multiple agent harnesses and learning methods, revealing limitations of current agents and providing insights for future design.
\end{itemize}

\FloatBarrier

\section{Related Work}
\label{sec:related}

\paragraph{Continual Learning Benchmarks.}
Evaluating continual learning capabilities has become an increasingly active topic for LLMs and agents~\citep{gaoSurveySelfEvolvingAgents2025,fangComprehensiveSurveySelfEvolving2025,zhengLifelongLearningLarge2025}.
Existing benchmarks span several distinct domains.
In knowledge-intensive tasks, prior work evaluates whether agents can effectively transfer procedural experience across coding, reasoning, and question-answering tasks~\citep{gaoEvoAgentBenchBenchmarkingAgent2026,wuStreamBenchBenchmarkingContinuous2024a,shuAgentCLRigorousEvaluation2026}.
In interactive digital environments, benchmarks test continual learning capabilities in web automation, operating systems, and knowledge graphs~\citep{jiangSEAEvalBenchmarkEvaluating2026,zhengLifelongAgentBenchEvaluatingLLM2025b}.
In long-term simulated environments, benchmarks model persistent virtual worlds where past behaviors continuously shape future states and decisions~\citep{caiBuildingSelfEvolvingAgents2025}.
Another major direction focuses on skill-centric evolution, studying whether agents can abstract reusable skills from experience and transfer them across tasks~\citep{zhongSkillLearnBenchBenchmarkingContinual2026,zhangSkillFlowBenchmarkingLifelong2026,guanContinualSkillBenchCanLLM2026,lei2026skillevolbench}.
Despite this progress, these benchmarks primarily operate in digital or symbolic environments, leaving continual learning in embodied scenarios largely unexplored.

\paragraph{Embodied AI Benchmarks.}
Embodied AI benchmarks evaluate the capabilities ranging from perception to long-horizon task execution~\citep{Liu_2025,yifanEmbodiedAISurvey2025,liuAligningCyberSpace2025}.
At the perception level, benchmarks test 3D scene understanding~\citep{wangEmbodiedScanHolisticMultiModal2024}, spatial reasoning~\citep{duEmbSpatialBenchBenchmarkingSpatial2024}, physical world understanding~\citep{chowPhysBenchBenchmarkingEnhancing2025}, and scene dynamics~\citep{wangEmbodiedSceneUnderstanding2025}.
For navigation, the field has progressed from language-conditioned navigation~\citep{andersonVisionandLanguageNavigationInterpreting2018,kuRoomAcrossRoomMultilingualVisionandLanguage2020a} to sound-guided navigation~\cite{liu2026omniecho}, object grounding~\citep{qiREVERIERemoteEmbodied2020}, multi-modal lifelong navigation~\citep{khannaGOATBenchBenchmarkMultiModal2024a}, and active exploration for question answering~\citep{majumdarOpenEQAEmbodiedQuestion2024}.
In manipulation, benchmarks evaluate multi-task generalization~\citep{yuMetaWorldBenchmarkEvaluation2020,jamesRLBenchRobotLearning2020}, object-level skill transfer~\citep{guManiSkill2UnifiedBenchmark2023a}, lifelong knowledge transfer~\citep{liuLIBEROBenchmarkingKnowledge2023}, and everyday task scalability~\citep{nasirianyRoboCasaLargeScaleSimulation2024}.
For long-horizon planning, benchmarks require agents to follow compositional instructions~\citep{shridharALFREDBenchmarkInterpreting2020}, interact through dialogue~\citep{padmakumarTEAChTaskdrivenEmbodied2021}, compose language-conditioned skills~\citep{meesCALVINBenchmarkLanguageConditioned2022}, handle diverse household activities~\citep{liBEHAVIOR1KBenchmarkEmbodied2023}, and adapt plans based on visual feedback~\citep{tupiniAsgardBenchEvaluatingVisually2026}.
These benchmarks comprehensively cover perception, navigation, manipulation, and planning.
Recent works further explore general-purpose embodied agent evaluation and realistic simulation environments~\citep{peng2024tong,ren2025simworld}.
However, none is designed for evaluating the continual learning capability of skills in long-horizon tasks.

\paragraph{Compositional Generalization.}
Compositional generalization, the ability to recombine learned elements for solving novel problems, has been studied across multiple domains.
In digital environments, benchmarks test whether mobile and web agents can compose atomic operations into multi-step tasks~\citep{guoAtomictoCompositionalGeneralizationMobile2025,furutaExposingLimitationsLanguage2023}.
In video understanding, benchmarks study whether models can recognize unseen action--object combinations from separately learned components~\citep{liC2CComponenttoCompositionLearning2024,ahnWhyCantOpen2026}.
For reasoning tasks, benchmarks show that models proficient in individual abilities still fail when these abilities must be combined~\citep{alazrakiAgentCoMaCompositionalBenchmark2026,liUnveilingCompositionalAbility2025a}.
In embodied manipulation, benchmarks evaluate whether robot policies can generalize to unseen skill compositions after training on atomic skills~\citep{wuATOMBenchRealWorldBenchmark2026,chenRoboHiManHierarchicalEvaluation2025,chenDeCoTaskDecomposition2025,hareshClevrSkillsCompositionalLanguage2024a}.
In contrast, \ours{} evaluates the continual learning capability in embodied simulation without any benchmark-specific training.

\section{Benchmark Construction}
\label{sec:method}

\begin{figure*}[t]
\centering
\includegraphics[width=\textwidth]{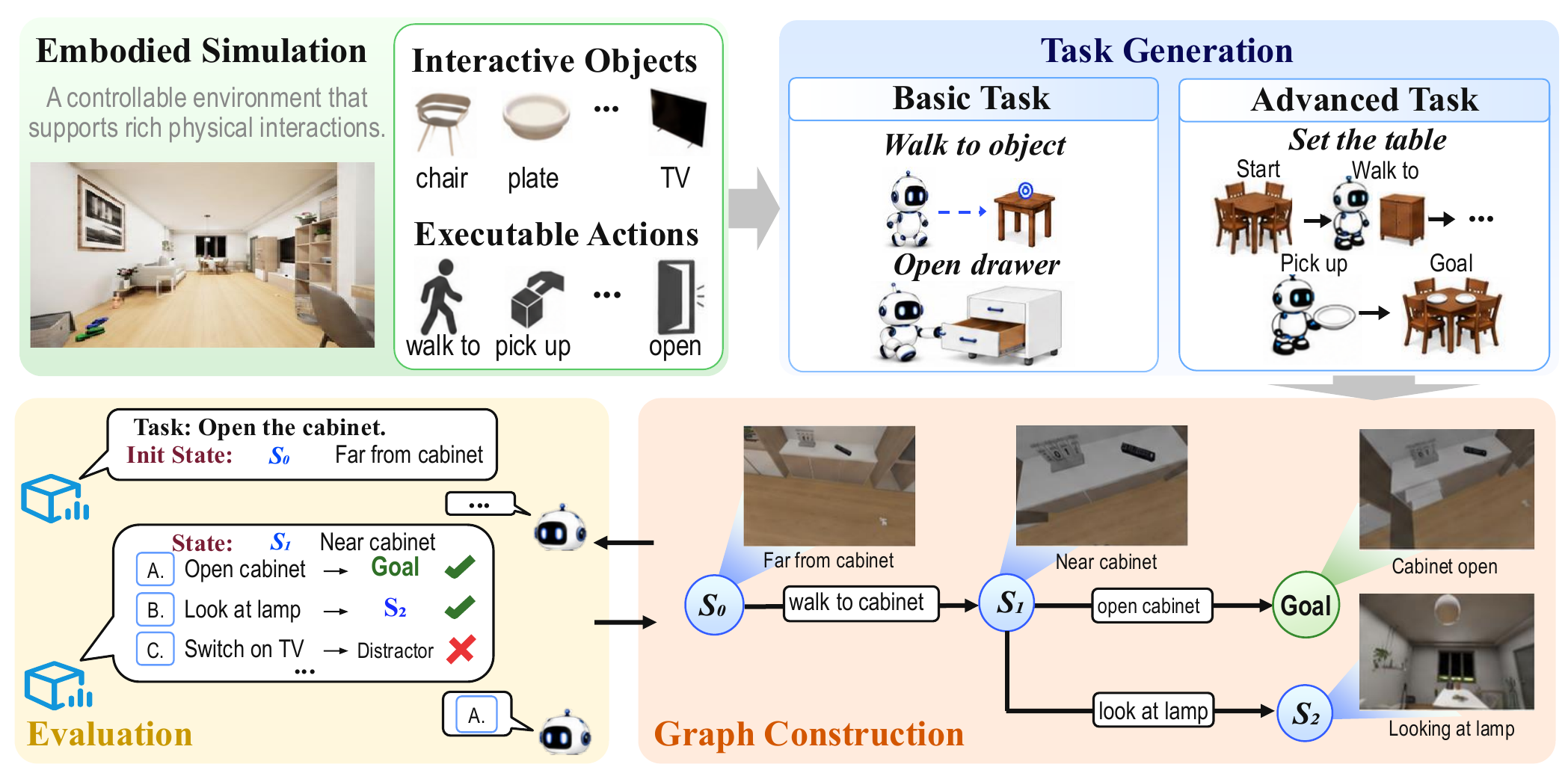}
\vspace{-8mm}
\caption{\textbf{Overview of the construction process of \ours{}.} During task generation, basic tasks connect a few atomic actions into simple household operations, and advanced tasks combine elements from multiple basic tasks into compositional tasks.
For each task, a state-transition graph is built, with each graph node as an intermediate state and paired with action options and a first-person-view image.
During evaluation, the agent observes the current state and selects from candidate actions. After each selection, the agent transitions to the next node on the graph.}
\vspace{-2mm}
\label{fig:pipeline}
\end{figure*}

As illustrated in Figure~\ref{fig:pipeline}, the construction of \ours{} consists of four stages: (1)~environment setup, (2)~task generation, (3)~graph construction, and (4)~evaluation protocol. We introduce each stage in the following subsections.

\subsection{Environment Setup: Embodied Simulation}
\label{sec:environment}

Evaluating compositional generalization requires a large combinatorial space of object--action pairs. 
We thus adopt TongSim as the simulation platform~\citep{sunTongSIMGeneralPlatform2025a}. It provides a controllable environment that supports rich physical interactions and scales to diverse task configurations. It was also used in recent embodied agent evaluation studies~\citep{heAutomaticCognitiveTask2026,zhaoNormActBenchmarkingEmbodied2026}.

Specifically, TongSim is a simulation with household scenarios and supports a wide range of interactive \textbf{objects} (such as cups, plates, cabinets, lamps, and appliances) and executable \textbf{actions} (such as walk to, look at, pick up, put down, open, close, and switch on, etc.). This simulation offers three key advantages: (1)~the platform supports a large number of object types and action primitives, providing sufficient combinatorial diversity; (2)~all scenes are indoor domestic environments, so that the tasks are grounded in meaningful, real-world household activities; and (3)~TongSim has not been widely used as training data for existing foundation models, making it suitable for evaluating the generalization capability of agents with reduced risk of data contamination.

\subsection{Task Generation: Compositional Generalization}
\label{sec:task}

Our goal is to evaluate the capability of continual learning. However, it is an abstract concept that is challenging to measure directly. We operationalize it through compositional generalization: if an agent has learned from basic tasks, it should recombine previously mastered skill elements to solve novel tasks rather than learning each new task from scratch. This leads to a natural evaluation protocol---let the agent first complete simpler tasks to accumulate experience, then test whether it can leverage that experience to solve more complex, unseen tasks.

We design two levels of tasks based on atomic actions available in TongSim.
Specifically, an \textbf{atomic action} is a single interaction with an object (e.g., ``pick up the cup''). It consists of a sequence of \textbf{steps}  (e.g., walking to the cup, looking at it, and picking it up), each of which is one executable action.
Building on it, a \textbf{basic task} composes a small number of atomic actions into a practical household operation (e.g., ``put the cup on the table'', which consists of ``pick up the cup'' and ``put down the cup on the table'').
Further, an \textbf{advanced task} composes more atomic actions than a basic task (e.g., ``tidy up the room'', which requires picking up the book, putting the book on the shelf, closing the cabinet, and switching off the lamp). We provide more details in the benchmark overview (Section~\ref{sec:setup}).
All task descriptions are generated by Qwen3.7-Plus and manually verified for correctness (see Appendix~\ref{app:construction_details}).

\subsection{Graph Construction: Action Trajectory Annotation}
\label{sec:graph}

For TongSim simulation, we chose its closed-source version, which provides more diverse actions than the open-source version. As a result, online simulation is not practical for the benchmark evaluation. Therefore, we construct an offline, reproducible evaluation process in the form of a state-transition graph, as shown in Figure~\ref{fig:pipeline}.
A graph structure naturally represents the task procedure: at each state, an agent selects an action and transitions to the next state. This graph structure offers three advantages: (1)~offline evaluation can be run in parallel without running the simulator; (2)~flexible control over the task complexity through graph branching and depth; and (3)~fine-grained analysis of agent behavior by checking each decision against the pre-computed graph.

Specifically, each graph node represents a unique environment state (e.g., ``near the desk and looking at the lamp'').
It is paired with an image captured from TongSim and a set of action candidates presented to the agent during evaluation.
These candidates are generated by Qwen3.7-Plus and manually verified for correctness.
Each executable candidate corresponds to an edge that deterministically transitions to the next state.
Given a task, we first define its initial and goal states and then use breadth-first search to find the shortest path between them.
Since each state supports only a bounded set of objects and actions, the reachable state space is compact enough for reliable graph search. We construct the distractor paths by expanding alternative executable actions from nodes along the shortest path. These paths are executable but deviate from the shortest path.
The detailed construction pipeline and task examples are provided in Appendices~\ref{app:construction_details} and~\ref{app:task_examples}.

\subsection{Evaluation Protocol}
\label{sec:eval_protocol}

With the task graphs built, we define how agents are evaluated on them.
The evaluation consists of a learning phase and a testing phase.
In the learning phase, the agent completes basic tasks to accumulate experience.
In the testing phase, the agent attempts unseen advanced tasks with this experience.
In this setup, basic tasks serve as the memory~\citep{xu2026contextual, huang2026m} and test-time learning data~\citep{sun2020test,liu2021ttt++} for the self-evolving agents.

For each task, the evaluation follows an interactive, multi-turn format:
(1)~the agent receives a task description and an image of the initial state;
(2)~at each step, the agent selects one action from a set of action candidates;
(3)~based on the action decision, the agent transitions to the next state and receives a new image with updated objects and action candidates;
(4)~this repeats until the agent reaches the goal (success) or exhausts the step budget (failure).
The action candidates at each step are designed into three types: optimal actions that follow the shortest path, suboptimal actions that are executable but deviate from the shortest path, and invalid actions that cannot be executed in the current state.
This interactive format preserves the sequential decision-making structure of embodied tasks while enabling fully offline, deterministic, and reproducible evaluation. 

\FloatBarrier

\section{Main Experiments}
\label{sec:exp}

In this section, we first describe the experimental setup, including the benchmark overview, evaluation settings, agent configurations, and metrics. We then report and discuss the main results.

\begin{figure*}[!b]
\centering
\vspace*{-5mm} 
\includegraphics[width=\textwidth]{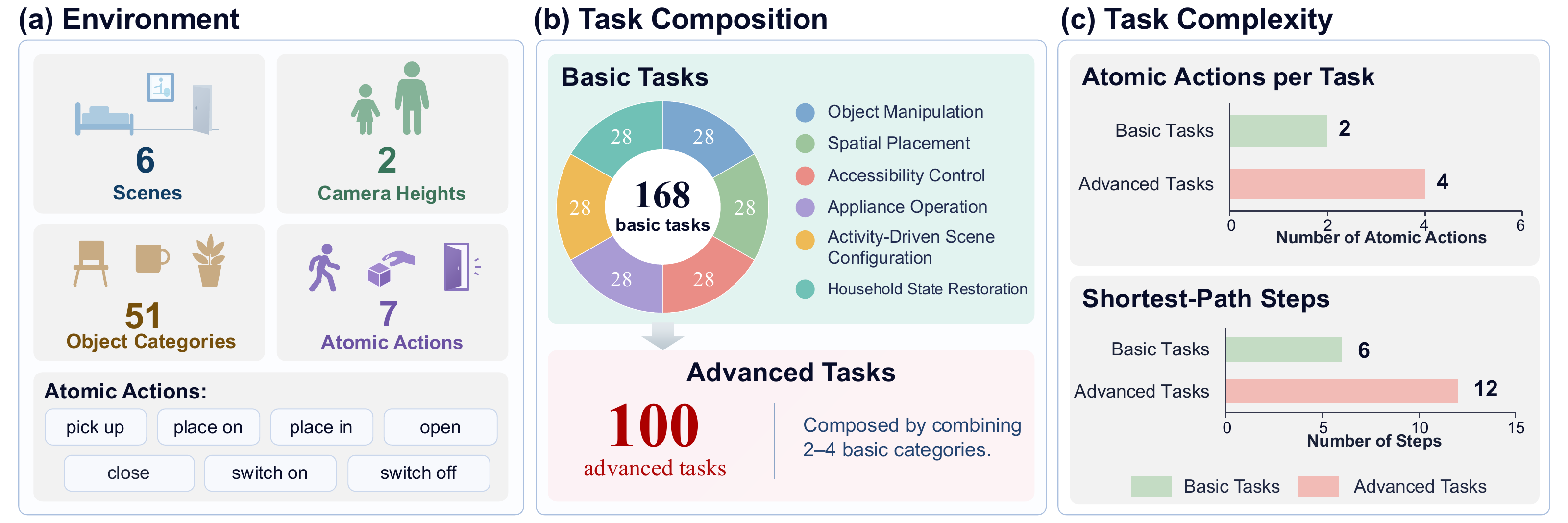}
\vspace{-8mm}
\caption{Benchmark statistics of \ours{}: environment configuration (a), task composition across six categories (b), and task complexity at two levels (c).}
\label{fig:stats}
\end{figure*}

\subsection{Experimental Setup}
\label{sec:setup}

\paragraph{Benchmark overview.}
Figure~\ref{fig:stats} summarizes the overall statistics of \ours{}.
The environment covers diverse scenes, object categories, and atomic actions, providing a broad combinatorial space for task construction.
The benchmark contains 268 tasks: 168 basic tasks and 100 advanced tasks.
Basic tasks span six categories, with each grounded in a distinct type of daily household activity.
Advanced tasks require substantially more atomic actions and shortest-path steps.
The atomic actions of each advanced task come from two to four categories of basic tasks.
Each atomic action consists of 3 steps.
Therefore, the shortest path has 6 steps for a basic task (2 atomic actions) and 12 steps for an advanced task (4 atomic actions).
With this scale, real-world grounding, and structured difficulty, \ours{} can reveal meaningful differences in the continual learning capability of agents.
Examples of basic and advanced tasks are provided in Appendix~\ref{app:task_examples}.

\paragraph{Evaluation settings.}
We evaluate agents under three settings:
(1)~\textbf{Zero-shot:} the agent performs basic and advanced tasks directly, without a learning phase.
(2)~\textbf{In-Context Learning (ICL):} in the learning phase, the agent completes two related basic tasks for each advanced task. In the testing phase, the full interaction trajectories of these basic tasks are retained in context.
(3)~\textbf{Skill Learning:} in the learning phase, the agent completes the same basic tasks as in ICL. It then writes a concise skill summary of its experience in Markdown format. In the testing phase, only this skill summary is provided in context, instead of the interaction trajectories.

The step budget of each task is the length of its shortest path plus extra steps. In the testing phase, we allow 3 extra steps as a small tolerance for suboptimal decisions. In the learning phase, we allow 6 extra steps for more exploration.

\paragraph{Agent configurations.}
We evaluate all tasks under each of four agent harnesses: \textsc{OpenClaw}~\citep{AgentHarnessPlugins2026}, \textsc{Claude Code}~\citep{HowClaudeCode2026}, \textsc{Codex}~\citep{OpenAICodexCLI2026}, and \textsc{Hermes}~\citep{HermesAgent2026}. We pair each harness with two backbone models through their official APIs: Qwen3.8-Flash~\citep{QwenAPIPlatform2026} and GPT5.6-Terra~\citep{GPT56TerraModelOpenAI2026}.

\begin{table}[H]
\centering
\caption{\textbf{Main results on \ours{}.}
For In-Context Learning and Skill Learning, agents first learn from two basic tasks and then attempt one advanced task.
}
\label{tab:main_results}
\vspace{0.35em}
\small
\setlength{\tabcolsep}{5pt}
\renewcommand{\arraystretch}{1.15}
\newcommand{\metriccell}[1]{\makebox[1.2cm][c]{#1}}
\begin{adjustbox}{max width=\linewidth}
\begin{tabular}{@{}ll cccc@{}}
\toprule
\multicolumn{2}{l}{\textbf{Success Rate (\%)}} &
\textsc{Basic Task} &
\multicolumn{3}{c}{\textsc{Advanced Task}} \\
\cmidrule(lr){3-3}\cmidrule(lr){4-6}
\textbf{Harness} & \textbf{Model} &
{Zero-shot} &
{Zero-shot} &
{In-Context Learning} &
{Skill Learning} \\
\midrule
\multirow{2}{*}{\textsc{Hermes}}
& Qwen3.8-Flash  & \metriccell{58.9} & \metriccell{8.0} & \metriccell{44.0} & \metriccell{33.0} \\
& GPT5.6-Terra     & \metriccell{53.0} & \metriccell{11.0} & \metriccell{16.0} & \metriccell{34.0} \\
\addlinespace[0.4em]
\multirow{2}{*}{\textsc{OpenClaw}}
& Qwen3.8-Flash  & \metriccell{50.6} & \metriccell{5.0} & \metriccell{35.0} & \metriccell{38.0} \\
& GPT5.6-Terra     & \metriccell{47.6} & \metriccell{12.0} & \metriccell{39.0} & \metriccell{22.0} \\
\addlinespace[0.4em]
\multirow{2}{*}{\textsc{Codex}}
& Qwen3.8-Flash  & \metriccell{52.4} & \metriccell{4.0} & \metriccell{34.0} & \metriccell{33.0} \\
& GPT5.6-Terra     & \metriccell{54.8} & \metriccell{11.0} & \metriccell{52.0} & \metriccell{26.0} \\
\addlinespace[0.4em]
\multirow{2}{*}{\textsc{Claude Code}}
& Qwen3.8-Flash  & \metriccell{42.9} & \metriccell{3.0} & \metriccell{27.0} & \metriccell{31.0} \\
& GPT5.6-Terra     & \metriccell{51.8} & \metriccell{9.0} & \metriccell{31.0} & \metriccell{25.0} \\
\bottomrule
\end{tabular}
\end{adjustbox}
\end{table}


\paragraph{Metrics.}
We report Success Rate (SR), which measures whether an agent can complete a task within the allotted step budget:
\begin{equation}
\mathrm{SR} = \frac{1}{N}\sum_{i=1}^{N} c_i\,,\quad c_i \in \{0,1\} \text{ indicates whether task } i \text{ is completed.}
\end{equation}
Success rate directly reflects continual learning capability---whether accumulated experience enables the agent to solve novel, compositionally harder tasks.

\subsection{Results}
\label{sec:main_results}

Table~\ref{tab:main_results} presents the main results.
Agents achieve a moderate zero-shot success rate on basic tasks (around 50\%), indicating that current agents can execute simple household tasks.
Zero-shot success rate on advanced tasks drops to around 10\%, validating that advanced tasks are genuinely difficult.
Both ICL and Skill Learning substantially improve advanced task success rate, raising it from around 10\% to above 30\%. This improvement is consistent across all harnesses, suggesting that experience from basic tasks can effectively transfer to unseen advanced tasks. 

\infobox{\textbf{Finding 1}: While advanced tasks are hardly solvable under zero-shot setting, experience from basic tasks enables consistent improvement across different harnesses and models.}

\paragraph{Discussion.} 
Between ICL and Skill Learning, there is no consistent advantage of one over the other.
We conjecture that both methods can extract meaningful information from the agent experience. 
Besides, the performance across harnesses is close yet far from saturation, suggesting great potential for improvement. Designing harnesses for continual learning can be a future direction.

\FloatBarrier

\section{Analysis}
\label{sec:analysis}
 
\subsection{Effects of Compositional Complexity}
\label{sec:complexity}
 
\paragraph{Motivation and Setting.}
Compositional generalization is the core approach for evaluating continual learning in \ours{}. A natural question follows: does more complex composition make continual learning harder? In the main experiment, advanced tasks consist of 4 atomic actions. Here, we extend our evaluation data to advanced tasks with 3 to 6 atomic actions to control compositional complexity. We measure both the raw success rate at each level and the gain over zero shot, defined as the success rate of In-Context Learning or Skill Learning minus the zero-shot success rate. We use \textsc{Hermes} with Qwen3.8-Flash for this analysis.

\vspace{-2mm}
\begin{figure}[H]
\centering
\includegraphics[width=\linewidth]{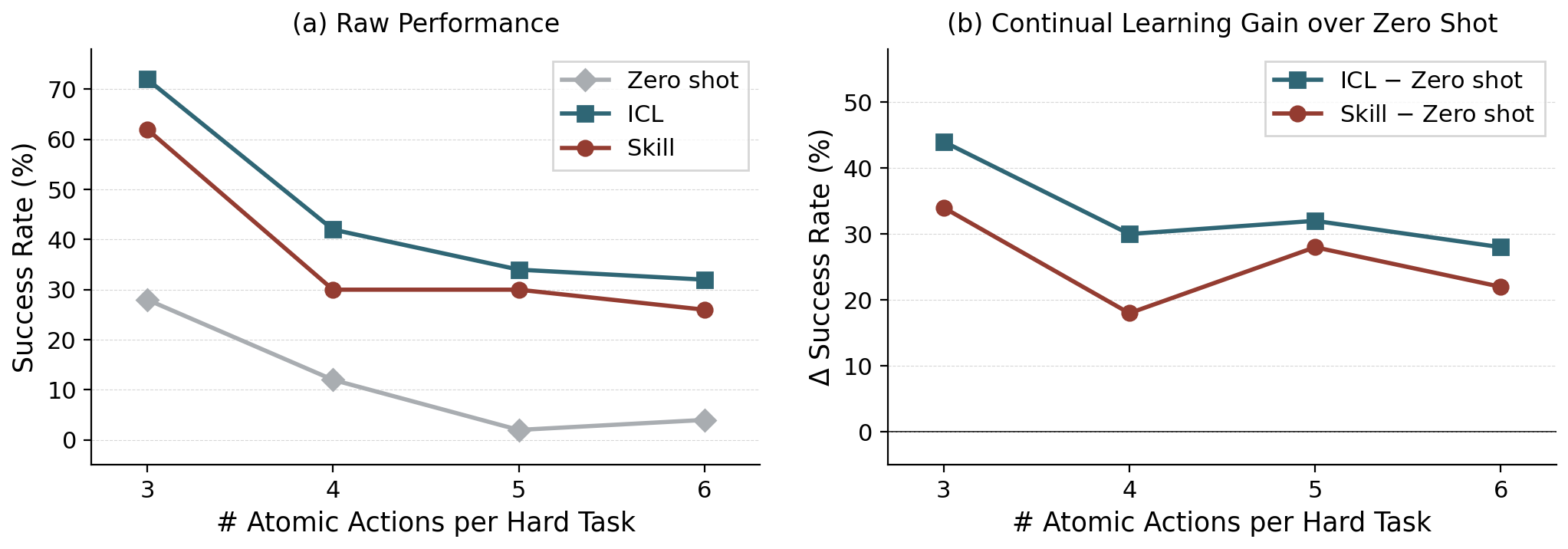}
\vspace{-8mm}
\caption{\textbf{Effect of compositional complexity on continual learning.} The $x$-axis varies the number of atomic actions per advanced task from 3 to 6. (a)~reports the raw success rate achieved by each setting. (b)~reports the gain over zero shot, defined as the success rate of ICL or Skill Learning minus that of zero shot, isolating the improvement attributed to the learning phase.}
\vspace{-4mm}
\label{fig:complexity}
\end{figure}

\paragraph{Results and Analysis.}
As shown in Figure~\ref{fig:complexity}(a), success rate decreases across all three settings as atomic actions increase from 3 to 6.
The largest drop occurs between 3 and 4 atomic actions (e.g., ICL falls from 72.0\% to 42.0\%), after which the success rate levels off.
In the zero-shot setting, success rate decreases sharply (e.g., from 28.0\% to 4.0\%).
In contrast, ICL and Skill Learning remain more stable in relative terms.
Specifically, Figure~\ref{fig:complexity}(b) shows that the gain over zero shot drops initially but then stabilizes at around 30\%.
This indicates that experience from basic tasks remains helpful as compositional complexity increases.
At the highest complexity (6 atomic actions), zero-shot success rate drops to 4.0\%, while ICL and Skill Learning still achieve 32.0\% and 26.0\%, respectively.
This shows that experience from basic tasks still brings a substantial gain over zero shot, even when zero-shot agents can barely solve the tasks.
 
\infobox{\textbf{Finding 2}: In zero-shot setting, success rate decreases sharply as complexity increases. In contrast, the gain over zero-shot setting remains stable, indicating that the benefit of continual learning is less sensitive to task difficulty.}
 
\subsection{Scaling with the Number of Learning Samples}
\label{sec:scaling}
 
\paragraph{Motivation and Setting.}
Continual learning assumes that accumulated experience improves an agent's performance on harder tasks. If this holds, more experience should lead to a higher success rate on advanced tasks. In the main experiment, agents learn from 2 basic tasks before attempting advanced tasks. Here, we increase the number of learning samples (i.e., basic tasks in the learning phase) from $k{=}0$ (zero shot) to $k{=}6$. We measure the success rate on the same set of advanced tasks under both In-Context Learning and Skill Learning settings. We evaluate two agent harnesses, \textsc{Hermes} and \textsc{OpenClaw}, both using Qwen3.8-Flash as the backbone model.
 
\begin{wrapfigure}[18]{l}{0.48\columnwidth}
\vspace{-5mm}
\centering
\includegraphics[width=\linewidth]{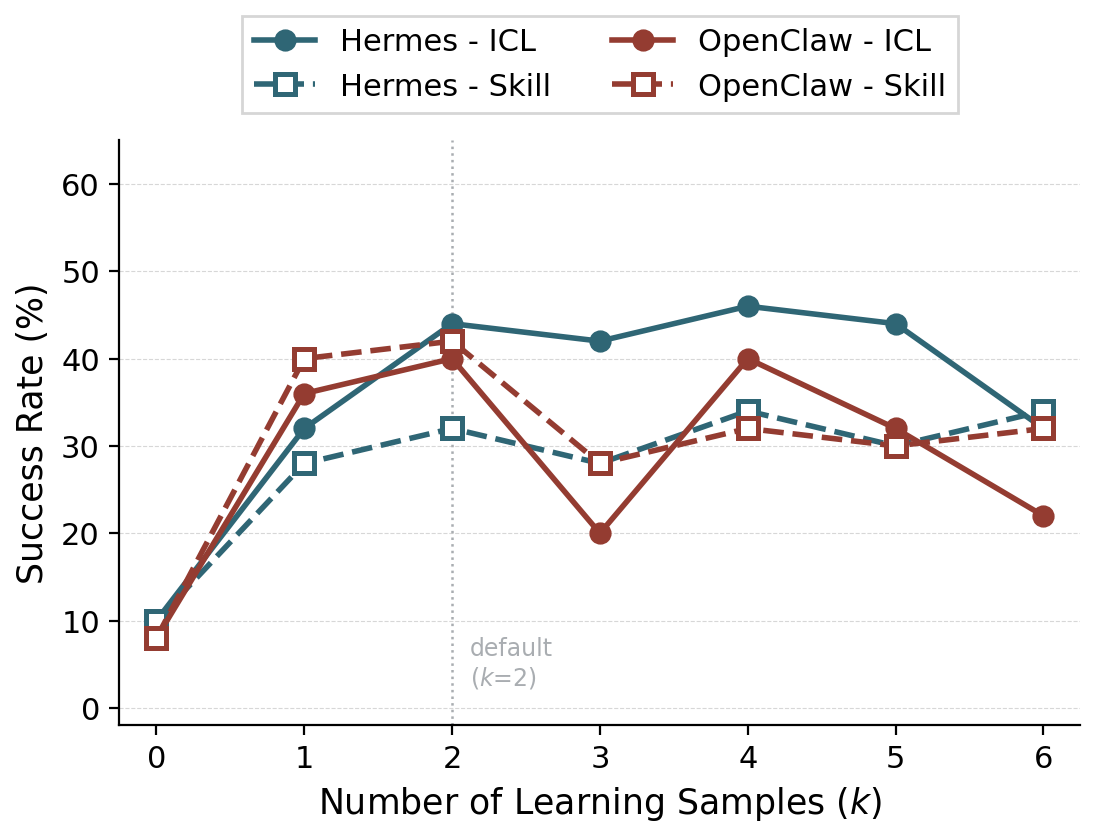}
\vspace{-8mm}
\caption{\textbf{Success rate versus the number of learning samples.} Each learning sample is a basic task that the agent completes before being tested on advanced tasks.}
\vspace{-4mm}
\label{fig:n_sample}
\end{wrapfigure}

\noindent\textbf{Results and Analysis.}
As shown in Figure~\ref{fig:n_sample}, the most substantial improvement occurs between $k{=}0$ and $k{=}1$.
A single learning sample raises the success rate from 8.0\%--10.0\% to 28.0\%--40.0\% across both harnesses and learning methods.
This initial jump accounts for most of the total improvement across the $k$ range.
Even minimal experience (completing just one basic task) provides sufficient new information for the agent to substantially improve on unseen advanced tasks.
Beyond $k{=}1$, however, success rate does not continue to rise but instead fluctuates without a clear upward trend.
This pattern remains across ICL and Skill Learning, and across both harnesses.
None of the four curves shows consistent improvement with additional learning samples.
This suggests that current agents cannot extract sufficiently distinct information from additional learning samples.
Therefore, scalable continual learning, where success rate keeps improving as learning samples increase, remains an open challenge for embodied agents.
 
\infobox{\textbf{Finding 3}: Success rate does not scale accordingly with the number of learning samples. A single learning sample significantly improves performance, while additional samples yield no consistent further gain.}

\begin{figure*}[!b]
\centering
\vspace{-6mm}
\includegraphics[width=0.92\textwidth]{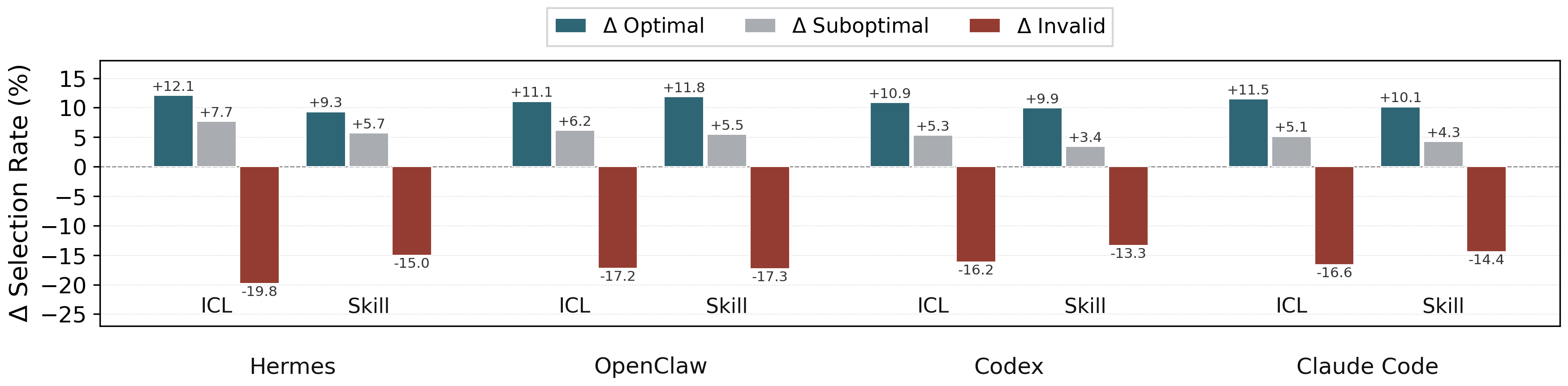}
\vspace{-4mm}
\caption{\textbf{Error analysis of decision behavior.}
Each group of bars shows the change in selection rate relative to the zero-shot setting for three action types:
\emph{optimal actions} follow the shortest path,
\emph{suboptimal actions} are executable but deviate from the shortest path,
and \emph{invalid actions} cannot be executed in the current state.}
\label{fig:action_analysis}
\end{figure*}
 
\subsection{Error Analysis}
\label{sec:action_analysis}

\paragraph{Motivation and Setting.}
The main results show that ICL and Skill Learning improve success rate, but do not reveal what agents actually learn from experience. As defined in Section~\ref{sec:eval_protocol}, the action candidates at each step fall into three types: optimal actions that follow the shortest path, suboptimal actions that are executable but deviate from the shortest path, and invalid actions that cannot be executed in the current state. We analyze how agents choose among these three action types. Instead of running new experiments, we use the selected actions recorded at each decision step in the main experiment (all four harnesses with Qwen3.8-Flash). Specifically, we compare the selection rate of each action type under ICL and Skill Learning with that in the zero-shot setting.

\noindent\textbf{Results and Analysis.}
As shown in Figure~\ref{fig:action_analysis}, experience sharply reduces invalid actions across all harnesses (by 13\%--20\%).
Fewer invalid actions allow agents to stay on executable trajectories and reach the goal within the step budget. This suggests that the reduction in invalid actions contributes to the higher success rates in Table~\ref{tab:main_results}.
Meanwhile, optimal actions increase substantially.
Suboptimal actions also increase slightly (by 3\%--8\%).
Overall, experience shifts the selections of agents from invalid actions to optimal and suboptimal actions.
ICL and Skill Learning show the same direction of change, but Skill Learning consistently produces smaller shifts across all harnesses.
In contrast, Table~\ref{tab:main_results} shows no consistent advantage between ICL and Skill Learning in success rate.
This may reflect a difference in how experience is represented.
Interaction trajectories may induce larger changes in decision behavior, while skill summaries may induce smaller but more focused changes.
 
\infobox{\textbf{Finding 4}: Continual learning alters the decision behavior of agents, resulting in an increase in optimal actions and a decrease in infeasible actions.}

\subsection{Step Budget Sensitivity}
\label{sec:budget}
 
\paragraph{Motivation and Setting.}
Each task has a shortest path to the goal, but agents often make suboptimal decisions along the way. As described in Section~\ref{sec:setup}, agents are allowed 3 extra steps by default as a small tolerance for suboptimal decisions. Here, we vary the number of extra steps from 0 to 6. We examine how the step budget affects success rate under different evaluation settings. We use \textsc{Hermes} with Qwen3.8-Flash for this analysis.

\begin{wrapfigure}[17]{l}{0.48\columnwidth}
\vspace{-5mm}
\centering
\includegraphics[width=\linewidth]{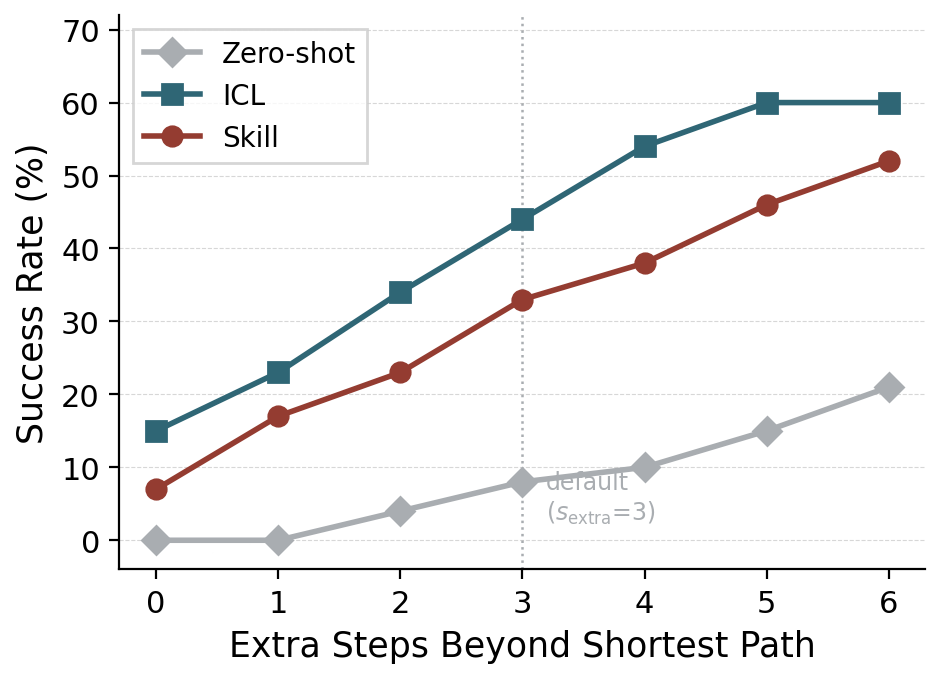}
\vspace{-8mm}
\caption{\textbf{Step budget sensitivity.} The $x$-axis shows the number of extra steps allowed beyond the shortest path. By default, agents are allowed 3 extra steps in the testing phase.}
\vspace{4mm}
\label{fig:step_budget}
\end{wrapfigure}

\noindent\textbf{Results and Analysis.}
As shown in Figure~\ref{fig:step_budget}, all three settings show a monotonically increasing success rate as the budget grows from 0 to 6 extra steps.
This is expected, since a larger budget gives agents more room for suboptimal decisions.
Across all budget values, the ranking of the three settings remains consistent.
ICL achieves the highest success rate, followed by Skill Learning and then zero shot.
This indicates that the gain over zero shot holds across different budgets.
At the strictest budget (0 extra steps), zero-shot success rate is 0.0\%, while ICL and Skill Learning still achieve 15.0\% and 7.0\%, respectively.
Even under this strictest budget, learning from experience remains effective.
Notably, compared to the zero-shot setting, ICL and Skill Learning exhibit a steeper increase in success rate as the budget grows. This indicates that the agents with prior experience can better benefit from additional tolerance. Meanwhile, the gap between experienced agents and zero-shot agents remains substantial even at the most generous budget (6 extra steps). This suggests that additional tolerance cannot replace the role of prior experience.
 
\infobox{\textbf{Finding 5}: As the number of extra steps increases, success rate improves across all settings, and continual learning methods yield a steeper increase compared to the zero-shot setting.}
 
\FloatBarrier
 
\section{Conclusion}
\label{sec:conclusion}

We introduce \ours{}, the first embodied simulation benchmark for evaluating the continual learning capability of self-evolving agents. Our key idea is to formulate continual learning through compositional generalization: rather than solving each new task from scratch, an agent should accumulate experience from basic tasks and recombine previously acquired skills to solve unseen, more complex tasks. Following this principle, \ours{} organizes tasks into atomic actions, basic tasks, and advanced tasks, providing a structured protocol for measuring whether embodied agents can acquire, retain, and transfer experience over time.

Extensive evaluations across multiple agent harnesses and learning methods reveal both promising progress and clear limitations: 
(1) Experience from basic tasks substantially improves performance on unseen advanced tasks over zero-shot execution, showing that current agents can transfer previously acquired knowledge to new compositions. 
(2) However, the performance of existing agents is far from saturation. Designing effective continual learning methods for embodied agents is a promising direction to explore.
(3) Further, the benefit of learning does not consistently increase with the amount of experience. How to enable the agents to accumulate and exploit additional interactions is still an open problem.
Overall, \ours{} provides a principled testbed for studying self-evolution in embodied agents and highlights the need for stronger mechanisms that can continuously consolidate, generalize, and reuse experience across tasks.
 
\bibliography{iclr2027_conference}

@article{liu2026omniecho,
	title={OmniEcho: Audio-Visual Spatial Understanding for Omni-Modal Embodied Agents}, 
	journal = {arXiv preprint arXiv:2609.23407},
	author={Ruixun Liu and Yuxuan Wang and Jiacheng Xie and Yuhuan You and Donghua Cai and Junming Lin and Xiong-Hui Chen and Zhifang Guo and Yunfei Chu and Qize Yang and Xize Cheng and Jin Xu and Yiwu Zhong},
	year = {2026},
}

@article{gaoSurveySelfEvolvingAgents2025,
	title = {A {Survey} of {Self}-{Evolving} {Agents}: {What}, {When}, {How}, and {Where} to {Evolve} on the {Path} to {Artificial} {Super} {Intelligence}},
	journal = {arXiv preprint arXiv:2507.21046},
	author = {Gao, Huan-ang and Geng, Jiayi and Hua, Wenyue and Hu, Mengkang and Juan, Xinzhe and Liu, Hongzhang and Liu, Shilong and Qiu, Jiahao and Qi, Xuan and Wu, Yiran and Wang, Hongru and Xiao, Han and Zhou, Yuhang and Zhang, Shaokun and Zhang, Jiayi and Xiang, Jinyu and Fang, Yixiong and Zhao, Qiwen and Liu, Dongrui and Ren, Qihan and Qian, Cheng and Wang, Zhenhailong and Hu, Minda and Wang, Huazheng and Wu, Qingyun and Ji, Heng and Wang, Mengdi},
	year = {2025},
}

@article{fangComprehensiveSurveySelfEvolving2025,
	title = {A {Comprehensive} {Survey} of {Self}-{Evolving} {AI} {Agents}: {A} {New} {Paradigm} {Bridging} {Foundation} {Models} and {Lifelong} {Agentic} {Systems}},
	journal = {arXiv preprint arXiv:2508.07407},
	author = {Fang, Jinyuan and Peng, Yanwen and Zhang, Xi and Wang, Yingxu and Yi, Xinhao and Zhang, Guibin and Xu, Yi and Wu, Bin and Liu, Siwei and Li, Zihao and Ren, Zhaochun and Aletras, Nikos and Wang, Xi and Zhou, Han and Meng, Zaiqiao},
	year = {2025},
}

@article{zhengLifelongLearningLarge2025,
	title = {Lifelong {Learning} of {Large} {Language} {Model} based {Agents}: {A} {Roadmap}},
	journal = {arXiv preprint arXiv:2501.07278},
	author = {Zheng, Junhao and Shi, Chengming and Cai, Xidi and Li, Qiuke and Zhang, Duzhen and Li, Chenxing and Yu, Dong and Ma, Qianli},
	year = {2025},
}

@article{gaoEvoAgentBenchBenchmarkingAgent2026,
	title = {{EvoAgentBench}: {Benchmarking} {Agent} {Self}-{Evolution} via {Ability} {Transfer}},
	journal = {arXiv preprint arXiv:2607.05202},
	author = {Gao, Xingze and Hu, Chuanrui and Chen, Hongda and Yao, Pengfei and Wang, Zhao and Bai, Yi and Wu, Zhengwei and Han, Yunyun and Cong, Xiaofeng and Gui, Jie and Deng, Yafeng and Li, Teng},
	year = {2026},
}

@article{wuStreamBenchBenchmarkingContinuous2024a,
	title = {{StreamBench}: {Towards} {Benchmarking} {Continuous} {Improvement} of {Language} {Agents}},
	journal = {arXiv preprint arXiv:2406.08747},
	author = {Wu, Cheng-Kuang and Tam, Zhi Rui and Lin, Chieh-Yen and Chen, Yun-Nung and Lee, Hung-yi},
	year = {2024},
}

@article{shuAgentCLRigorousEvaluation2026,
	title = {{AgentCL}: {Toward} {Rigorous} {Evaluation} of {Continual} {Learning} in {Language} {Agents}},
	journal = {arXiv preprint arXiv:2606.02461},
	author = {Shu, Yiheng and Gutiérrez, Bernal Jiménez and Jonnalagedda, Saisri Padmaja and Yao, Yuguang and Sun, Huan and Su, Yu},
	year = {2026},
}

@article{jiangSEAEvalBenchmarkEvaluating2026,
	title = {{SEA}-{Eval}: {A} {Benchmark} for {Evaluating} {Self}-{Evolving} {Agents} {Beyond} {Episodic} {Assessment}},
	journal = {arXiv preprint arXiv:2604.08988},
	author = {Jiang, Sihang and Ma, Lipeng and Hong, Zhonghua and Wang, Keyi and Lu, Zhiyu and Wang, Tengfei and Chen, Shisong and Zhang, Jinghao and Pan, Tianjun and Li, Weijia and Liang, Jiaqing and Xiao, Yanghua},
	year = {2026},
}

@article{zhengLifelongAgentBenchEvaluatingLLM2025b,
	title = {{LifelongAgentBench}: {Evaluating} {LLM} {Agents} as {Lifelong} {Learners}},
	journal = {arXiv preprint arXiv:2505.11942},
	author = {Zheng, Junhao and Cai, Xidi and Li, Qiuke and Zhang, Duzhen and Li, ZhongZhi and Zhang, Yingying and Song, Le and Ma, Qianli},
	year = {2025},
}

@article{caiBuildingSelfEvolvingAgents2025,
	title = {Building {Self}-{Evolving} {Agents} via {Experience}-{Driven} {Lifelong} {Learning}: {A} {Framework} and {Benchmark}},
	journal = {arXiv preprint arXiv:2508.19005},
	author = {Cai, Yuxuan and Hao, Yipeng and Zhou, Jie and Yan, Hang and Lei, Zhikai and Zhen, Rui and Han, Zhenhua and Yang, Yutao and Li, Junsong and Pan, Qianjun and Huai, Tianyu and Chen, Qin and Li, Xin and Chen, Kai and Zhang, Bo and Qiu, Xipeng and He, Liang},
	year = {2025},
}

@article{zhongSkillLearnBenchBenchmarkingContinual2026,
	title = {{SkillLearnBench}: {Benchmarking} {Continual} {Learning} {Methods} for {Agent} {Skill} {Generation} on {Real}-{World} {Tasks}},
	journal = {arXiv preprint arXiv:2604.20087},
	author = {Zhong, Shanshan and Lu, Yi and Ning, Jingjie and Wan, Yibing and Feng, Lihan and Ao, Yuyi and Ribeiro, Leonardo F. R. and Dreyer, Markus and Ammirati, Sean and Xiong, Chenyan},
	year = {2026},
}

@article{zhangSkillFlowBenchmarkingLifelong2026,
	title = {{SkillFlow}:{Benchmarking} {Lifelong} {Skill} {Discovery} and {Evolution} for {Autonomous} {Agents}},
	journal = {arXiv preprint arXiv:2604.17308},
	author = {Zhang, Ziao and Shi, Kou and Huang, Shiting and Nie, Avery and Zeng, Yu and Zhao, Yiming and Fang, Zhen and Su, Qishen and Qiu, Haibo and Yang, Wei and Ren, Qingnan and Zou, Shun and Huang, Wenxuan and Chen, Lin and Chen, Zehui and Zhao, Feng},
	year = {2026},
}

@article{guanContinualSkillBenchCanLLM2026,
	title = {{ContinualSkillBench}: {Can} {LLM} {Agents} {Truly} {Evolve} {Their} {Capabilities}?},
	journal = {arXiv preprint arXiv:2608.03874},
	author = {Guan, Tianyi and Wang, Yiding and Yang, Haotong and Cao, Siyuan and Liu, Shirui and Hu, Yi and Li, Jiaqi and Zhang, Muhan},
	year = {2026},
}

@article{Liu_2025,
	title = {Embodied Intelligence: A Synergy of Morphology, Action, Perception and Learning},
	author = {Liu, Huaping and Guo, Di and Cangelosi, Angelo},
	year = {2025},
	month = mar,
	journal = {ACM Computing Surveys},
	volume = {57},
	number = {7},
	pages = {1--36},
	publisher = {Association for Computing Machinery (ACM)},
	issn = {1557-7341},
	doi = {10.1145/3717059}
}

@article{yifanEmbodiedAISurvey2025,
	title = {Embodied {AI}: {A} {Survey} on the {Evolution} from {Perceptive} to {Behavioral} {Intelligence}},
	doi = {10.1002/smb2.70003},
	journal = {SmartBot},
	author = {Yifan, Chen and Wei, Mingjie and Wang, Xuesong and Liu, Yuanxing and Wang, Jizhe and Song, Hao and Ma, Longxuan and Di, Donglin and Sun, Churui and Liu, Kaifeng and Qi, Le and Yu, Jianing and Tian, Xiang and Liang, Shiqi and Duan, Chuanxi and Hong, Zijian and Zhang, Weinan and Liu, Ting},
	month = sep,
	year = {2025},
}

@article{liuAligningCyberSpace2025,
	title = {Aligning {Cyber} {Space} {With} {Physical} {World}: {A} {Comprehensive} {Survey} on {Embodied} {AI}},
	doi = {10.1109/tmech.2025.3574943},
	journal = {IEEE/ASME Transactions on Mechatronics},
	author = {Liu, Yang and Chen, Weixing and Bai, Yongjie and Liang, Xiaodan and Li, Guanbin and Gao, Wen and Lin, Liang},
	month = dec,
	year = {2025},
}

@inproceedings{wangEmbodiedScanHolisticMultiModal2024,
	title = {{EmbodiedScan}: {A} {Holistic} {Multi}-{Modal} {3D} {Perception} {Suite} {Towards} {Embodied} {AI}},
	booktitle = {Proceedings of the {IEEE}/{CVF} {Conference} on {Computer} {Vision} and {Pattern} {Recognition} ({CVPR})},
	author = {Wang, Tai and Mao, Xiaohan and Zhu, Chenming and Xu, Runsen and Lyu, Ruiyuan and Li, Peisen and Chen, Xiao and Zhang, Wenwei and Chen, Kai and Xue, Tianfan and Liu, Xihui and Lu, Cewu and Lin, Dahua and Pang, Jiangmiao},
	year = {2024},
	pages = {19757--19767},
}

@inproceedings{duEmbSpatialBenchBenchmarkingSpatial2024,
	title = {{EmbSpatial}-{Bench}: {Benchmarking} {Spatial} {Understanding} for {Embodied} {Tasks} with {Large} {Vision}-{Language} {Models}},
	doi = {10.18653/v1/2024.acl-short.33},
	booktitle = {Proceedings of the 62nd {Annual} {Meeting} of the {Association} for {Computational} {Linguistics} ({Volume} 2: {Short} {Papers})},
	publisher = {Association for Computational Linguistics},
	author = {Du, Mengfei and Wu, Binhao and Li, Zejun and Huang, Xuanjing and Wei, Zhongyu},
	year = {2024},
	pages = {346--355},
}

@article{chowPhysBenchBenchmarkingEnhancing2025,
	title = {{PhysBench}: {Benchmarking} and {Enhancing} {Vision}-{Language} {Models} for {Physical} {World} {Understanding}},
	journal = {arXiv preprint arXiv:2501.16411},
	author = {Chow, Wei and Mao, Jiageng and Li, Boyi and Seita, Daniel and Guizilini, Vitor and Wang, Yue},
	year = {2025},
}

@inproceedings{wangEmbodiedSceneUnderstanding2025,
	title = {Embodied {Scene} {Understanding} for {Vision} {Language} {Models} via {MetaVQA}},
	booktitle = {Proceedings of the {IEEE}/{CVF} {Conference} on {Computer} {Vision} and {Pattern} {Recognition} ({CVPR})},
	author = {Wang, Weizhen and Duan, Chenda and Peng, Zhenghao and Liu, Yuxin and Zhou, Bolei},
	year = {2025},
	pages = {22453--22464},
}

@inproceedings{andersonVisionandLanguageNavigationInterpreting2018,
	title = {Vision-and-{Language} {Navigation}: {Interpreting} {Visually}-{Grounded} {Navigation} {Instructions} in {Real} {Environments}},
	booktitle = {Proceedings of the {IEEE} {Conference} on {Computer} {Vision} and {Pattern} {Recognition} ({CVPR})},
	author = {Anderson, Peter and Wu, Qi and Teney, Damien and Bruce, Jake and Johnson, Mark and Sünderhauf, Niko and Reid, Ian and Gould, Stephen and van den Hengel, Anton},
	year = {2018},
}

@inproceedings{kuRoomAcrossRoomMultilingualVisionandLanguage2020a,
	title = {Room-{Across}-{Room}: {Multilingual} {Vision}-and-{Language} {Navigation} with {Dense} {Spatiotemporal} {Grounding}},
	doi = {10.18653/v1/2020.emnlp-main.356},
	booktitle = {Proceedings of the 2020 {Conference} on {Empirical} {Methods} in {Natural} {Language} {Processing} ({EMNLP})},
	publisher = {Association for Computational Linguistics},
	author = {Ku, Alexander and Anderson, Peter and Patel, Roma and Ie, Eugene and Baldridge, Jason},
	year = {2020},
	pages = {4392--4412},
}

@inproceedings{qiREVERIERemoteEmbodied2020,
	title = {{REVERIE}: {Remote} {Embodied} {Visual} {Referring} {Expression} in {Real} {Indoor} {Environments}},
	booktitle = {Proceedings of the {IEEE}/{CVF} {Conference} on {Computer} {Vision} and {Pattern} {Recognition} ({CVPR})},
	author = {Qi, Yuankai and Wu, Qi and Anderson, Peter and Wang, Xin and Wang, William Yang and Shen, Chunhua and Hengel, Anton van den},
	year = {2020},
}

@inproceedings{khannaGOATBenchBenchmarkMultiModal2024a,
	title = {{GOAT}-{Bench}: {A} {Benchmark} for {Multi}-{Modal} {Lifelong} {Navigation}},
	booktitle = {Proceedings of the {IEEE}/{CVF} {Conference} on {Computer} {Vision} and {Pattern} {Recognition} ({CVPR})},
	author = {Khanna, Mukul and Ramrakhya, Ram and Chhablani, Gunjan and Yenamandra, Sriram and Gervet, Theophile and Chang, Matthew and Kira, Zsolt and Chaplot, Devendra Singh and Batra, Dhruv and Mottaghi, Roozbeh},
	year = {2024},
	pages = {16373--16383},
}

@inproceedings{majumdarOpenEQAEmbodiedQuestion2024,
	title = {{OpenEQA}: {Embodied} {Question} {Answering} in the {Era} of {Foundation} {Models}},
	booktitle = {Proceedings of the {IEEE}/{CVF} {Conference} on {Computer} {Vision} and {Pattern} {Recognition} ({CVPR})},
	author = {Majumdar, Arjun and Ajay, Anurag and Zhang, Xiaohan and Putta, Pranav and Yenamandra, Sriram and Henaff, Mikael and Silwal, Sneha and Mcvay, Paul and Maksymets, Oleksandr and Arnaud, Sergio and Yadav, Karmesh and Li, Qiyang and Newman, Ben and Sharma, Mohit and Berges, Vincent and Zhang, Shiqi and Agrawal, Pulkit and Bisk, Yonatan and Batra, Dhruv and Kalakrishnan, Mrinal and Meier, Franziska and Paxton, Chris and Sax, Alexander and Rajeswaran, Aravind},
	year = {2024},
	pages = {16488--16498},
}

@inproceedings{yuMetaWorldBenchmarkEvaluation2020,
	series = {Proceedings of {Machine} {Learning} {Research}},
	title = {Meta-{World}: {A} {Benchmark} and {Evaluation} for {Multi}-{Task} and {Meta} {Reinforcement} {Learning}},
	volume = {100},
	booktitle = {Proceedings of the {Conference} on {Robot} {Learning}},
	publisher = {PMLR},
	author = {Yu, Tianhe and Quillen, Deirdre and He, Zhanpeng and Julian, Ryan and Hausman, Karol and Finn, Chelsea and Levine, Sergey},
	year = {2020},
	note = {Month: 30 Oct--01 Nov},
	pages = {1094--1100},
}

@article{jamesRLBenchRobotLearning2020,
	title = {{RLBench}: {The} {Robot} {Learning} {Benchmark} \& {Learning} {Environment}},
	volume = {5},
	doi = {10.1109/lra.2020.2974707},
	number = {2},
	journal = {IEEE Robotics and Automation Letters},
	author = {James, Stephen and Ma, Zicong and Arrojo, David Rovick and Davison, Andrew J.},
	month = apr,
	year = {2020},
	pages = {3019--3026},
}

@inproceedings{guManiSkill2UnifiedBenchmark2023a,
	title = {{ManiSkill2}: {A} {Unified} {Benchmark} for {Generalizable} {Manipulation} {Skills}},
	booktitle = {International {Conference} on {Learning} {Representations}},
	author = {Gu, Jiayuan and Xiang, Fanbo and Li, Xuanlin and Ling, Zhan and Liu, Xiqiang and Mu, Tongzhou and Tang, Yihe and Tao, Stone and Wei, Xinyue and Yao, Yunchao and Yuan, Xiaodi and Xie, Pengwei and Huang, Zhiao and Chen, Rui and Su, Hao},
	year = {2023},
}

@article{liuLIBEROBenchmarkingKnowledge2023,
	title = {{LIBERO}: {Benchmarking} {Knowledge} {Transfer} for {Lifelong} {Robot} {Learning}},
	journal = {arXiv preprint arXiv:2306.03310},
	author = {Liu, Bo and Zhu, Yifeng and Gao, Chongkai and Feng, Yihao and Liu, Qiang and Zhu, Yuke and Stone, Peter},
	year = {2023},
}

@article{nasirianyRoboCasaLargeScaleSimulation2024,
	title = {{RoboCasa}: {Large}-{Scale} {Simulation} of {Everyday} {Tasks} for {Generalist} {Robots}},
	journal = {arXiv preprint arXiv:2406.02523},
	author = {Nasiriany, Soroush and Maddukuri, Abhiram and Zhang, Lance and Parikh, Adeet and Lo, Aaron and Joshi, Abhishek and Mandlekar, Ajay and Zhu, Yuke},
	year = {2024},
}

@inproceedings{shridharALFREDBenchmarkInterpreting2020,
	title = {{ALFRED}: {A} {Benchmark} for {Interpreting} {Grounded} {Instructions} for {Everyday} {Tasks}},
	booktitle = {Proceedings of the {IEEE}/{CVF} {Conference} on {Computer} {Vision} and {Pattern} {Recognition} ({CVPR})},
	author = {Shridhar, Mohit and Thomason, Jesse and Gordon, Daniel and Bisk, Yonatan and Han, Winson and Mottaghi, Roozbeh and Zettlemoyer, Luke and Fox, Dieter},
	year = {2020},
}

@article{padmakumarTEAChTaskdrivenEmbodied2021,
	title = {{TEACh}: {Task}-driven {Embodied} {Agents} that {Chat}},
	journal = {arXiv preprint arXiv:2110.00534},
	author = {Padmakumar, Aishwarya and Thomason, Jesse and Shrivastava, Ayush and Lange, Patrick and Narayan-Chen, Anjali and Gella, Spandana and Piramuthu, Robinson and Tur, Gokhan and Hakkani-Tur, Dilek},
	year = {2021},
}

@article{meesCALVINBenchmarkLanguageConditioned2022,
	title = {{CALVIN}: {A} {Benchmark} for {Language}-{Conditioned} {Policy} {Learning} for {Long}-{Horizon} {Robot} {Manipulation} {Tasks}},
	volume = {7},
	doi = {10.1109/lra.2022.3180108},
	number = {3},
	journal = {IEEE Robotics and Automation Letters},
	author = {Mees, Oier and Hermann, Lukas and Rosete-Beas, Erick and Burgard, Wolfram},
	month = jul,
	year = {2022},
	pages = {7327--7334},
}

@inproceedings{liBEHAVIOR1KBenchmarkEmbodied2023,
	series = {Proceedings of {Machine} {Learning} {Research}},
	title = {{BEHAVIOR}-{1K}: {A} {Benchmark} for {Embodied} {AI} with 1,000 {Everyday} {Activities} and {Realistic} {Simulation}},
	volume = {205},
	booktitle = {Proceedings of {The} 6th {Conference} on {Robot} {Learning}},
	publisher = {PMLR},
	author = {Li, Chengshu and Zhang, Ruohan and Wong, Josiah and Gokmen, Cem and Srivastava, Sanjana and Mart{\'\i}n-Mart{\'\i}n, Roberto and Wang, Chen and Levine, Gabrael and Lingelbach, Michael and Sun, Jiankai and Anvari, Mona and Hwang, Minjune and Sharma, Manasi and Aydin, Arman and Bansal, Dhruva and Hunter, Samuel and Kim, Kyu-Young and Lou, Alan and Matthews, Caleb R and Villa-Renteria, Ivan and Tang, Jerry Huayang and Tang, Claire and Xia, Fei and Savarese, Silvio and Gweon, Hyowon and Liu, Karen and Wu, Jiajun and Fei-Fei, Li},
	year = {2023},
	note = {Month: 14--18 Dec},
	pages = {80--93},
}

@article{tupiniAsgardBenchEvaluatingVisually2026,
	title = {{AsgardBench} -- {Evaluating} {Visually} {Grounded} {Interactive} {Planning} {Under} {Minimal} {Feedback}},
	journal = {arXiv preprint arXiv:2603.15888},
	author = {Tupini, Andrea and Liden, Lars and Tan, Reuben and Wang, Yu and Gao, Jianfeng},
	year = {2026},
}

@article{wuATOMBenchRealWorldBenchmark2026,
	title = {{ATOM}-{Bench}: {A} {Real}-{World} {Benchmark} for {Atomic} {Skills} and {Compositional} {Generalization} in {Manipulation} {Policies}},
	journal = {arXiv preprint arXiv:2606.16826},
	author = {Wu, Zenan and Wei, Bingqing and Liu, Lu and He, Zheqi and Wang, Xi and Liu, Jiakang and Li, Zehui and Yao, Guocai and Zheng, Jing-Shu and Yang, Xi and Wang, Yongtao},
	year = {2026},
}

@article{chenRoboHiManHierarchicalEvaluation2025,
	title = {{RoboHiMan}: {A} {Hierarchical} {Evaluation} {Paradigm} for {Compositional} {Generalization} in {Long}-{Horizon} {Manipulation}},
	journal = {arXiv preprint arXiv:2510.13149},
	author = {Chen, Yangtao and Chen, Zixuan and Chan, Nga Teng and Chen, Junting and Yin, Junhui and Shi, Jieqi and Gao, Yang and Li, Yong-Lu and Huo, Jing},
	year = {2025},
}

@article{chenDeCoTaskDecomposition2025,
	title = {{DeCo}: {Task} {Decomposition} and {Skill} {Composition} for {Zero}-{Shot} {Generalization} in {Long}-{Horizon} {3D} {Manipulation}},
	journal = {arXiv preprint arXiv:2505.00527},
	author = {Chen, Zixuan and Yin, Junhui and Chen, Yangtao and Huo, Jing and Tian, Pinzhuo and Shi, Jieqi and Hou, Yiwen and Li, Yinchuan and Gao, Yang},
	year = {2025},
}

@inproceedings{hareshClevrSkillsCompositionalLanguage2024a,
	title = {{ClevrSkills}: {Compositional} {Language} {And} {Visual} {Reasoning} in {Robotics}},
	volume = {37},
	doi = {10.52202/079017-1209},
	booktitle = {Advances in {Neural} {Information} {Processing} {Systems}},
	author = {Haresh, Sanjay and Dijkman, Daniel and Bhattacharyya, Apratim and Memisevic, Roland},
	year = {2024},
	pages = {38235--38266},
}

@article{guoAtomictoCompositionalGeneralizationMobile2025,
	title = {Atomic-to-{Compositional} {Generalization} for {Mobile} {Agents} with {A} {New} {Benchmark} and {Scheduling} {System}},
	journal = {arXiv preprint arXiv:2506.08972},
	author = {Guo, Yuan and Miao, Tingjia and Wu, Zheng and Cheng, Pengzhou and Zhou, Ming and Zhang, Zhuosheng},
	year = {2025},
}

@article{furutaExposingLimitationsLanguage2023,
	title = {Exposing {Limitations} of {Language} {Model} {Agents} in {Sequential}-{Task} {Compositions} on the {Web}},
	journal = {arXiv preprint arXiv:2311.18751},
	author = {Furuta, Hiroki and Matsuo, Yutaka and Faust, Aleksandra and Gur, Izzeddin},
	year = {2023},
}

@inproceedings{alazrakiAgentCoMaCompositionalBenchmark2026,
  title={Agentcoma: A compositional benchmark mixing commonsense and mathematical reasoning in real-world scenarios},
  author={Alazraki, Lisa and Chen, Lihu and Brassard, Ana and Stacey, Joe and Rahmani, Hossein A and Rei, Marek},
  booktitle={Proceedings of the 64th Annual Meeting of the Association for Computational Linguistics (Volume 1: Long Papers)},
  pages={8383--8410},
  year={2026}
}

@inproceedings{liUnveilingCompositionalAbility2025a,
	title = {Unveiling the {Compositional} {Ability} {Gap} in {Vision}-{Language} {Reasoning} {Model}},
	volume = {38, Main Conference},
	doi = {10.52202/085713-4222},
	booktitle = {Advances in {Neural} {Information} {Processing} {Systems}},
	publisher = {Curran Associates, Inc.},
	author = {Li, Tianle and Zhang, Jihai and Rao, Yongming and Cheng, Yu},
	year = {2025},
	pages = {126574--126592},
}

@article{liC2CComponenttoCompositionLearning2024,
	title = {{C2C}: {Component}-to-{Composition} {Learning} for {Zero}-{Shot} {Compositional} {Action} {Recognition}},
	journal = {arXiv preprint arXiv:2407.06113},
	author = {Li, Rongchang and Feng, Zhenhua and Xu, Tianyang and Li, Linze and Wu, Xiao-Jun and Awais, Muhammad and Atito, Sara and Kittler, Josef},
	year = {2024},
}

@misc{ahnWhyCantOpen2026,
	title = {Why {Can}'t {I} {Open} {My} {Drawer}? {Mitigating} {Object}-{Driven} {Shortcuts} in {Zero}-{Shot} {Compositional} {Action} {Recognition}},
	shorttitle = {Why {Can}'t {I} {Open} {My} {Drawer}?},
	doi = {10.48550/arXiv.2601.16211},
	language = {en-US},
	urldate = {2026-07-11},
	publisher = {arXiv},
	author = {Ahn, Geo and Lee, Inwoong and Kim, Taeoh and Shim, Minho and Wee, Dongyoon and Choi, Jinwoo},
	month = jul,
	year = {2026},
	note = {arXiv:2601.16211 [cs.CV]},
}

@article{sunTongSIMGeneralPlatform2025a,
	title = {{TongSIM}: {A} {General} {Platform} for {Simulating} {Intelligent} {Machines}},
	journal = {arXiv preprint arXiv:2512.20206},
	author = {Sun, Zhe and Wu, Kunlun and Fu, Chuanjian and Song, Zeming and Shi, Langyong and Xue, Zihe and Jing, Bohan and Yang, Ying and Gao, Xiaomeng and Li, Aijia and Guo, Tianyu and Li, Huiying and Yang, Xueyuan and Liu, Rongkai and He, Xinyi and Wang, Yuxi and Li, Yue and Liu, Mingyuan and Lu, Yujie and Xie, Hongzhao and Zhao, Shiyun and Dai, Bo and Wang, Wei and Yuan, Tao and Zhu, Song-Chun and Peng, Yujia and Zhang, Zhenliang},
	year = {2025},
}

@article{heAutomaticCognitiveTask2026,
	title = {Automatic {Cognitive} {Task} {Generation} for {In}-{Situ} {Evaluation} of {Embodied} {Agents}},
	journal = {arXiv preprint arXiv:2602.05249},
	author = {He, Xinyi and Yang, Ying and Fu, Chuanjian and Guo, Sihan and Zhu, Songchun and Fan, Lifeng and Zhang, Zhenliang and Peng, Yujia},
	year = {2026},
}

@article{zhaoNormActBenchmarkingEmbodied2026,
	title = {{NormAct}: {Benchmarking} {Embodied} {Agents}' {Proactive} {Compliance} with {Unspoken} {Social} {Norms}},
	journal = {arXiv preprint arXiv:2606.27826},
	author = {Zhao, Shiyun and Song, Xinwei and Guo, Tianyu and Gao, Xiaomeng and Liu, Mingyuan and Han, Xu and Zhang, Yuanyuan and Zhang, Zhenliang and Feng, Xue and Dai, Bo},
	year = {2026},
}

@misc{AgentHarnessPlugins2026,
	title = {{OpenClaw} --- Open-Source {AI} Assistant},
	author = {{OpenClaw}},
	url = {https://openclaw.ai/},
	year = {2026},
}

@misc{HowClaudeCode2026,
	title = {{Claude} {Code}},
	author = {{Anthropic}},
	url = {https://claude.com/product/claude-code},
	year = {2026},
}

@misc{OpenAICodexCLI2026,
	title = {{OpenAI} {Codex} {CLI} - {Getting} {Started}},
	author = {{OpenAI}},
	url = {https://help.openai.com/en/articles/11096431},
	year = {2026},
}

@misc{HermesAgent2026,
	title = {Hermes {Agent}},
	author = {{Nous Research}},
	url = {https://github.com/NousResearch/hermes-agent},
	year = {2026},
}

@misc{QwenAPIPlatform2026,
	title = {{Qwen3.8-Flash}},
	author = {{Qwen}},
	url = {https://www.qwencloud.com/models/qwen3.8-flash},
	year = {2026},
}

@misc{GPT56TerraModelOpenAI2026,
	title = {{GPT}-5.6 {Terra} {Model} {\textbar} {OpenAI} {API}},
	author = {{OpenAI}},
	url = {https://developers.openai.com/api/docs/models/gpt-5.6-terra},
	year = {2026},
}

@inproceedings{sun2020test,
  title={Test-time training with self-supervision for generalization under distribution shifts},
  author={Sun, Yu and Wang, Xiaolong and Liu, Zhuang and Miller, John and Efros, Alexei and Hardt, Moritz},
  booktitle={International conference on machine learning},
  pages={9229--9248},
  year={2020},
  organization={PMLR}
}

@article{liu2021ttt++,
  title={Ttt++: When does self-supervised test-time training fail or thrive?},
  author={Liu, Yuejiang and Kothari, Parth and Van Delft, Bastien and Bellot-Gurlet, Baptiste and Mordan, Taylor and Alahi, Alexandre},
  journal={Advances in Neural Information Processing Systems},
  volume={34},
  pages={21808--21820},
  year={2021}
}

@article{lei2026skillevolbench,
  title={SkillEvolBench: Benchmarking the Evolution from Episodic Experience to Procedural Skills},
  author={Lei, Yingtie and Wan, Zhongwei and Zhang, Jiankun and Alam, Samiul and Zhong, Zixuan and Huang, Peizhou and Wang, Xin and Zhang, Jingxuan and Zhou, Donghao and Hsieh, Yunta and others},
  journal={arXiv preprint arXiv:2605.24117},
  year={2026}
}

@article{peng2024tong,
  title={The tong test: Evaluating artificial general intelligence through dynamic embodied physical and social interactions},
  author={Peng, Yujia and Han, Jiaheng and Zhang, Zhenliang and Fan, Lifeng and Liu, Tengyu and Qi, Siyuan and Feng, Xue and Ma, Yuxi and Wang, Yizhou and Zhu, Song-Chun},
  journal={Engineering},
  volume={34},
  pages={12--22},
  year={2024},
  publisher={Elsevier}
}

@article{ren2025simworld,
  title={Simworld: An open-ended realistic simulator for autonomous agents in physical and social worlds},
  author={Ren, Jiawei and Zhuang, Yan and Ye, Xiaokang and Mao, Lingjun and He, Xuhong and Shen, Jianzhi and Dogra, Mrinaal and Liang, Yiming and Zhang, Ruixuan and Yue, Tianai and others},
  journal={arXiv preprint arXiv:2512.01078},
  year={2025}
}

@article{xu2026contextual,
  title={Contextual agentic memory is a memo, not true memory},
  author={Xu, Binyan and Dai, Xilin and Zhang, Kehuan},
  journal={arXiv preprint arXiv:2604.27707},
  year={2026}
}

@article{huang2026m,
  title={M$^3$Eval: Multi-Modal Memory Evaluation through Cognitively-Grounded Video Tasks},
  author={Huang, Jie and Liu, Ruixun and Sun, Sirui and Yang, Xinyi and Li, Yin and Zhu, Yixin and Zhong, Yiwu},
  journal={arXiv preprint arXiv:2606.05008},
  year={2026}
}
\bibliographystyle{iclr2027_conference}
 
\clearpage
\appendix
\section{Basic Task Categories}
\label{app:task_categories}

As described in Section~\ref{sec:task}, basic tasks in \ours{} span six capability categories. Table~\ref{tab:task_categories} lists each category and its core evaluation question.

\begin{table}[h]
\centering
\caption{Six capability categories of basic tasks in \ours{}.}
\label{tab:task_categories}
\small
\begin{tabular}{@{}p{4.8cm}p{8.4cm}@{}}
\toprule
\textbf{Category} & \textbf{Core Question} \\
\midrule
Object Manipulation & Can the agent locate and manipulate everyday objects into a usable or manageable state? \\
\addlinespace
Spatial Placement & Can the agent understand spatial relationships between objects and surfaces or containers? \\
\addlinespace
Accessibility Control & Can the agent change the state of doors, drawers, and cabinet doors to enable or restrict access? \\
\addlinespace
Appliance Operation & Can the agent toggle household appliances and environmental devices? \\
\addlinespace
Activity-Driven Scene Configuration & Can the agent set up multiple objects and devices to support activities such as reading, dining, or leisure? \\
\addlinespace
Household State Restoration & Can the agent restore a room to a tidy, energy-efficient, and reusable state? \\
\bottomrule
\end{tabular}
\end{table}

\section{Graph Construction Pipeline}
\label{app:construction_details}

As described in Sections~\ref{sec:task} and~\ref{sec:eval_protocol}, \ours{} builds tasks from atomic actions available in TongSim. Given the atomic actions for each task, we first construct the state-transition graph. We then generate two types of textual content: task descriptions and per-state action options.

\paragraph{LLM Generation.}
Both task descriptions and action options are produced by a three-stage pipeline powered by Qwen3.7-Plus. The pipeline serves two purposes: converting low-level atomic actions into natural household scenarios, and ensuring that action options do not give away the correct answer through shortcuts. The key prompt specifications for all three stages are provided in Section~\ref{app:prompts}.

\paragraph{Human Verification.}
Three annotators independently verify each generated sample. The verification covers three aspects: captured images, task descriptions, and per-state action options. For each aspect, annotators check for factual errors, logical inconsistencies, and whether the content reflects realistic household activities. A sample is retained only if all three annotators approve it.

\subsection{Prompt Specifications}
\label{app:prompts}

\begin{promptbox}[Stage 1: Task Description Rewrite]
\begin{Verbatim}[fontsize=\scriptsize,breaklines=true,breakanywhere=true,breaksymbolleft={},breaksymbolright={},breakanywheresymbolpre={},breakanywheresymbolpost={}]
# Stage 1: Task Description Rewrite

## Objective

Rewrite a baseline task description into a natural, scene-based household request. The rewritten description is what the tested model actually sees. You do not change the underlying task structure -- only the surface expression.

## Task Hierarchy

- L1 (atomic goal): the smallest unit of household goal, e.g., "turn off the TV", "open the cabinet", "place the cup on the table".
- L2 (basic task): composed of 2 L1 goals forming a coherent household activity.
- L4 (advanced task): composed of 4 L1 goals combining multiple capability categories.

## Input Fields

- level: L2 or L4.
- ability_categories: capability categories involved (e.g., appliance operation, spatial placement).
- l1_goals: list of constituent L1 atomic goals with descriptions, action types, and target objects.
- baseline_task_description: a direct concatenation of all L1 goals (e.g., "Turn off the TV, open the lamp, close the cabinet door").

## Rewrite Principles

1. Merge multiple L1 goals into a single coherent living scenario (e.g., "A guest is arriving soon -- make the living room presentable and quiet"). The model should need to reason about what to do from the scene image, not read the steps directly from the text.

2. Do NOT produce a checklist, enumerate target objects by name, expose execution order, or translate L1 goals one-by-one with synonyms.

3. Replace specific object names with functional or contextual references (e.g., TV -> "noise source"; table lamp -> "reading light"; cabinet -> "storage area").

4. Add a plausible living scenario (guest arriving, bedtime routine, post-cooking cleanup, etc.) that makes all L1 goals co-exist naturally.

## Output

A single JSON object with fields: description_highlevel (the rewritten task description), rationale (why the rewrite is natural and leak-free), and anti_leak_check (boolean flags for exposes_order, is_checklist, mentions_specific_objects).
\end{Verbatim}
\end{promptbox}

\begin{promptbox}[Stage 2: Per-State Option Rewrite]
\begin{Verbatim}[fontsize=\scriptsize,breaklines=true,breakanywhere=true,breaksymbolleft={},breaksymbolright={},breakanywheresymbolpre={},breakanywheresymbolpost={}]
# Stage 2: Per-State Option Rewrite

## Objective

Rewrite the four hard-coded action options at a single graph state into natural language. After rewriting, all four options should look like plausible next steps, forcing the model to rely on the scene image and current state rather than linguistic cues.

## Underlying Execution Pattern (must be hidden)

Each household goal typically requires three phases:
1. Observe: look at the target object (look_at).
2. Approach: move to the target object (walk_to).
3. Operate: execute the action (open / close / switch_on / etc.).
If option wording reveals which phase it belongs to (e.g., observe options always say "check the status of X"), the model can guess the answer without seeing the image. The core challenge is to make all options sound like purposeful household actions regardless of their underlying phase.

## Option Roles

- success_path: the optimal next action. Must not stand out among the four options.
- valid_detour: executable and task-relevant, but not optimal. Must read like active task progress, not like a detour.
- invalid: cannot be executed in the current state. Must read like a plausible next step; its appeal comes from semantic relevance to the task, not from low-quality wording.

## Rewrite Rules

1. Each option: 1 sentence, 10-25 English words, naming the involved object or area and connecting to the high-level task scenario.
2. Two-layer fit: every option (including distractors) must fit both the current sub-goal and the high-level task description.
3. Hide phases: do not let observe options read as "status checks", approach options as "moving closer to operate", or operate options as the only ones with a sense of completion.
4. Uniform style: all four options should be similar in length, detail, and directness. The correct option must not be the only one containing task keywords.
5. Invalid options must not contain words like "wrong", "invalid", or "not needed", and must not explain why the action cannot be executed.

## Output

A single JSON object containing: state_id, options (each with id, text, role, design_note), and self_check (correct_stands_out, current_goal_fit, highlevel_task_fit, phase_exposed, has_implementation_terms).
\end{Verbatim}
\end{promptbox}

\begin{promptbox}[Stage 3: Cross-State Option Hardening]
\begin{Verbatim}[fontsize=\scriptsize,breaklines=true,breakanywhere=true,breaksymbolleft={},breaksymbolright={},breakanywheresymbolpre={},breakanywheresymbolpost={}]
# Stage 3: Cross-State Option Hardening

## Objective

Given a batch of states from the same task graph, detect and eliminate cross-state linguistic patterns that the tested model could exploit. The model sees its full interaction history, so any recurring pattern across consecutive steps can be discovered and used to bypass visual reasoning.

## Context

Each household goal takes three phases (observe -> approach -> operate). A basic task has 2 such goals, producing a repeating 2-beat rhythm on the shortest path:
Goal 1: observe -> approach -> operate
Goal 2: observe -> approach -> operate
Stage 2 made each state's options internally deceptive but could not address cross-state patterns. This stage fixes that.

## Input

A batch of states (either the full shortest path or a branch cluster), each containing: state_id, state_context, belongs_to_l1 (which sub-goal this state belongs to), and options (with baseline_text, rewritten_text, role, action_type, object).

## Patterns to Detect and Eliminate

1. Three-beat rhythm: correct options (success_path) across consecutive states show a cyclic "attention/focus -> head/move -> execute/switch" pattern that repeats per sub-goal.
2. Style bias: correct options are systematically shorter, more direct, or contain more task keywords than distractors.
3. Expression repetition: the same phrasing or sentence structure appears across multiple states.

## Hardening Strategies

A. Blur phase boundaries: make observe-phase correct options sound like preparation or positioning rather than looking. Make approach-phase correct options sound like executing part of an action rather than walking.
B. Mix phase styles in distractors: in an observe state, give some distractors an operate-style wording; in an operate state, give some distractors an observe-style wording.
C. Break repetition across sub-goals: if sub-goal 1's correct options follow pattern X, sub-goal 2's correct options must use a different pattern.
D. Vary detail level: some correct options should be short and direct, others should include contextual detail, matching the variation in distractors.
E. Vary verbs and sentence structures: the same action type appearing in different states must use different verbs and phrasings.

## Constraints

- Modify rewritten_text only (not state_id, id, role, action_type, or baseline_text).
- Edits are adjustments to Stage 2 output, not full rewrites from baseline_text.
- Options must remain specific, natural, and connected to the high-level task scenario.
- Do not use implementation terms (graph, state, precondition, phase).

## Output

A single JSON object containing: states (each with state_id, modified flag, options, modification_note) and pattern_report (documenting which patterns were found and how they were fixed).
\end{Verbatim}
\end{promptbox}

\section{Task Examples}
\label{app:task_examples}

\begin{figure}[H]
\centering
\begin{minipage}[t]{0.24\textwidth}
\centering
\includegraphics[width=\linewidth]{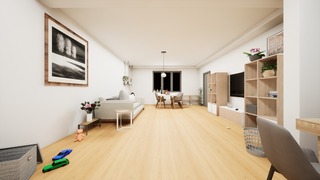}
\par\scriptsize State 0
\end{minipage}\hfill
\begin{minipage}[t]{0.24\textwidth}
\centering
\includegraphics[width=\linewidth]{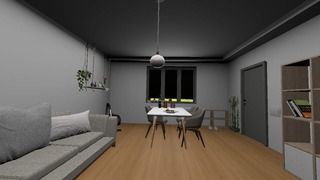}
\par\scriptsize State 1
\end{minipage}\hfill
\begin{minipage}[t]{0.24\textwidth}
\centering
\includegraphics[width=\linewidth]{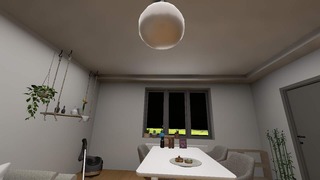}
\par\scriptsize State 2
\end{minipage}\hfill
\begin{minipage}[t]{0.24\textwidth}
\centering
\includegraphics[width=\linewidth]{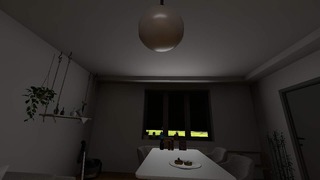}
\par\scriptsize State 3
\end{minipage}
\par\vspace{0.5em}
\begin{minipage}[t]{0.24\textwidth}
\centering
\includegraphics[width=\linewidth]{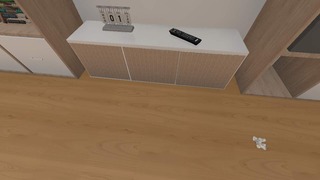}
\par\scriptsize State 4
\end{minipage}\hfill
\begin{minipage}[t]{0.24\textwidth}
\centering
\includegraphics[width=\linewidth]{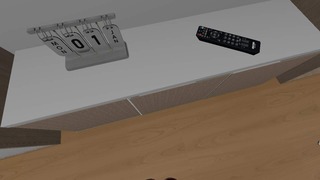}
\par\scriptsize State 5
\end{minipage}\hfill
\begin{minipage}[t]{0.24\textwidth}
\centering
\includegraphics[width=\linewidth]{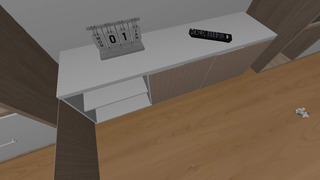}
\par\scriptsize State 6
\end{minipage}\hfill\makebox[0.24\textwidth]{}
\vspace{-2mm}
\caption{Visual states along the shortest path for a basic-task example.}
\label{fig:tongbench-example-states}
\end{figure}

\begin{figure}[H]
\centering
\begin{minipage}[t]{0.24\textwidth}
\centering
\includegraphics[width=\linewidth]{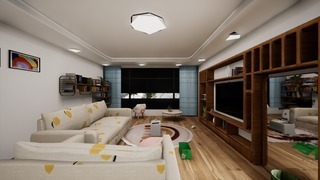}
\par\scriptsize State 0
\end{minipage}\hfill
\begin{minipage}[t]{0.24\textwidth}
\centering
\includegraphics[width=\linewidth]{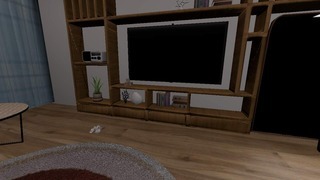}
\par\scriptsize State 1
\end{minipage}\hfill
\begin{minipage}[t]{0.24\textwidth}
\centering
\includegraphics[width=\linewidth]{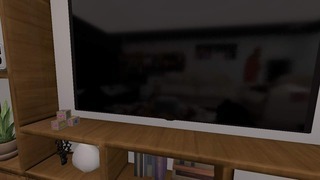}
\par\scriptsize State 2
\end{minipage}\hfill
\begin{minipage}[t]{0.24\textwidth}
\centering
\includegraphics[width=\linewidth]{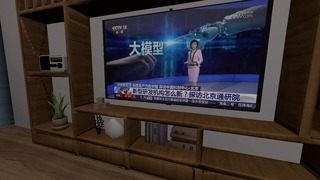}
\par\scriptsize State 3
\end{minipage}
\par\vspace{0.5em}
\begin{minipage}[t]{0.24\textwidth}
\centering
\includegraphics[width=\linewidth]{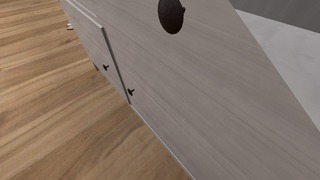}
\par\scriptsize State 4
\end{minipage}\hfill
\begin{minipage}[t]{0.24\textwidth}
\centering
\includegraphics[width=\linewidth]{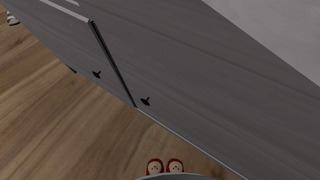}
\par\scriptsize State 5
\end{minipage}\hfill
\begin{minipage}[t]{0.24\textwidth}
\centering
\includegraphics[width=\linewidth]{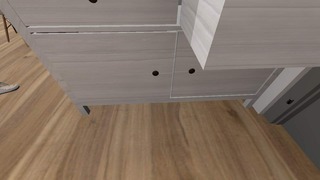}
\par\scriptsize State 6
\end{minipage}\hfill\makebox[0.24\textwidth]{}
\vspace{-2mm}
\caption{Visual states along the shortest path for a basic-task example.}
\label{fig:tongbench-l2-example-0022-states}
\end{figure}

\begin{figure}[H]
\centering
\begin{minipage}[t]{0.24\textwidth}
\centering
\includegraphics[width=\linewidth]{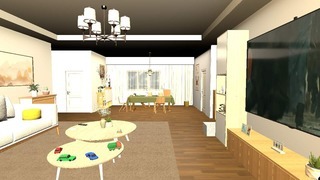}
\par\scriptsize State 0
\end{minipage}\hfill
\begin{minipage}[t]{0.24\textwidth}
\centering
\includegraphics[width=\linewidth]{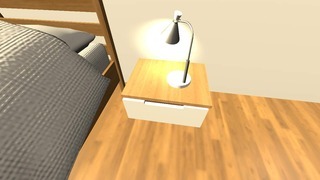}
\par\scriptsize State 1
\end{minipage}\hfill
\begin{minipage}[t]{0.24\textwidth}
\centering
\includegraphics[width=\linewidth]{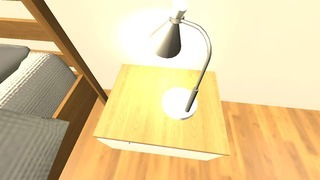}
\par\scriptsize State 2
\end{minipage}\hfill
\begin{minipage}[t]{0.24\textwidth}
\centering
\includegraphics[width=\linewidth]{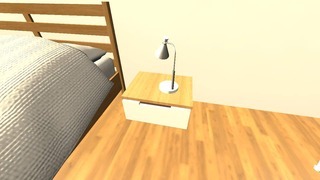}
\par\scriptsize State 3
\end{minipage}
\par\vspace{0.5em}
\begin{minipage}[t]{0.24\textwidth}
\centering
\includegraphics[width=\linewidth]{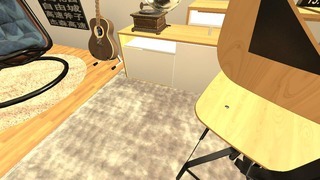}
\par\scriptsize State 4
\end{minipage}\hfill
\begin{minipage}[t]{0.24\textwidth}
\centering
\includegraphics[width=\linewidth]{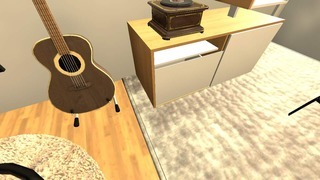}
\par\scriptsize State 5
\end{minipage}\hfill
\begin{minipage}[t]{0.24\textwidth}
\centering
\includegraphics[width=\linewidth]{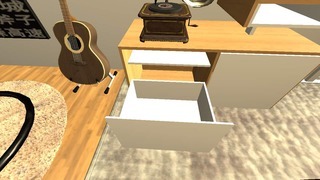}
\par\scriptsize State 6
\end{minipage}\hfill
\begin{minipage}[t]{0.24\textwidth}
\centering
\includegraphics[width=\linewidth]{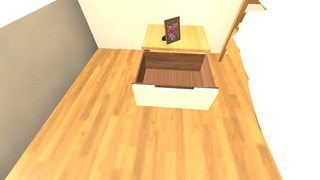}
\par\scriptsize State 7
\end{minipage}
\par\vspace{0.5em}
\begin{minipage}[t]{0.24\textwidth}
\centering
\includegraphics[width=\linewidth]{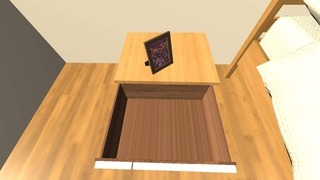}
\par\scriptsize State 8
\end{minipage}\hfill
\begin{minipage}[t]{0.24\textwidth}
\centering
\includegraphics[width=\linewidth]{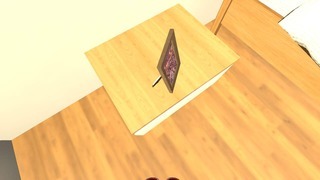}
\par\scriptsize State 9
\end{minipage}\hfill
\begin{minipage}[t]{0.24\textwidth}
\centering
\includegraphics[width=\linewidth]{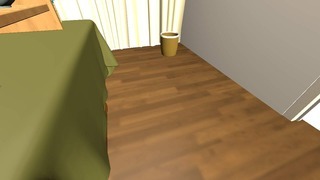}
\par\scriptsize State 10
\end{minipage}\hfill
\begin{minipage}[t]{0.24\textwidth}
\centering
\includegraphics[width=\linewidth]{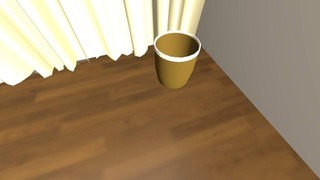}
\par\scriptsize State 11
\end{minipage}
\par\vspace{0.5em}
\begin{minipage}[t]{0.24\textwidth}
\centering
\includegraphics[width=\linewidth]{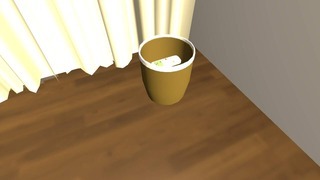}
\par\scriptsize State 12
\end{minipage}
\vspace{-2mm}
\caption{Visual states along the shortest path for an advanced-task example, including the final state.}
\label{fig:tongbench-l4-example-states}
\end{figure}

We provide task examples at both compositional levels to illustrate the evaluation format. The agent first receives a natural-language task description. At each step, it observes the current-state image and selects one action from multiple candidates. Specifically, we include two basic-task examples (Figures~\ref{fig:tongbench-example-states} and~\ref{fig:tongbench-l2-example-0022-states}; Tables~\ref{tab:tongbench-example-options} and~\ref{tab:tongbench-l2-example-0022-options}) and one advanced-task example (Figure~\ref{fig:tongbench-l4-example-states}; Table~\ref{tab:tongbench-l4-example-options}).

\raggedbottom
\setlength{\intextsep}{10pt}

\begingroup
\setlength{\tabcolsep}{3pt}
\renewcommand{\arraystretch}{1.15}
\setlength{\LTcapwidth}{\textwidth}
\setlength{\LTpre}{0pt}
\setlength{\LTpost}{1em}
\begin{longtable}{@{}>{\scriptsize\raggedright\arraybackslash}p{\dimexpr0.095\textwidth-3pt\relax}>{\scriptsize\raggedright\arraybackslash}p{\dimexpr0.08\textwidth-3pt\relax}>{\scriptsize\raggedright\arraybackslash}p{\dimexpr0.585\textwidth-6pt\relax}>{\scriptsize\raggedright\arraybackslash}p{\dimexpr0.24\textwidth-6pt\relax}@{}}
\caption{Basic-task example: action options and hidden graph roles at each state.}
\vspace{-2mm}
\label{tab:tongbench-example-options} \\
\toprule
\multicolumn{4}{@{}p{\textwidth}@{}}{\scriptsize \textbf{Task:} I'm about to watch a movie in the living room. Turn off the overhead light so the screen isn't washed out, and open up the media cabinet so everything's within reach.} \\
\midrule
\textbf{State} & \textbf{Option} & \textbf{Text shown to the agent} & \textbf{Hidden role} \\
\midrule
\endfirsthead
\multicolumn{4}{@{}l@{}}{\normalsize Table~\thetable\ (continued).} \\[0.5em]
\toprule
\textbf{State} & \textbf{Option} & \textbf{Text shown to the agent} & \textbf{Hidden role} \\
\midrule
\endhead
\midrule
\multicolumn{4}{@{}r@{}}{\scriptsize\textit{Continued on the next page.}} \\
\endfoot
\bottomrule
\endlastfoot
State 0 & A & Head over to the ceiling light area to get ready to turn it off. & Invalid in current state \\*
State 0 & B & Switch on the overhead light to brighten the living room before the movie. & Invalid in current state \\*
State 0 & C & Shut off the ceiling light right away to set the mood for the movie. & Invalid in current state \\*
State 0 & D & The overhead light is still on --- that's the first thing to deal with before the movie. & Correct shortest-path action \\
\midrule
State 1 & A & Check the ceiling light to see if it's already been switched off. & Invalid in current state \\*
State 1 & B & Switch off the ceiling light to darken the room for movie viewing. & Invalid in current state \\*
State 1 & C & Turn on the ceiling light to brighten the space before settling in for the movie. & Invalid in current state \\*
State 1 & D & Walk over to the ceiling light so you can reach the switch. & Correct shortest-path action \\
\midrule
State 2 & A & Switch off the overhead light --- the room needs to be dark for the movie. & Correct shortest-path action \\*
State 2 & B & Step closer to the ceiling fixture to get a better grip on the switch. & Executable detour \\*
State 2 & C & Glance up at the ceiling light to confirm how bright it still is. & Invalid in current state \\*
State 2 & D & Check the television area to see if the media setup is ready. & Executable detour \\
\midrule
State 3 & A & Head over to the television area to get the media setup going. & Invalid in current state \\*
State 3 & B & Walk toward the ceiling light to start working on the room's ambiance. & Executable detour \\*
State 3 & C & Switch on the overhead light to brighten things up before the movie. & Executable detour \\*
State 3 & D & With the light off, the media cabinet by the TV is the next thing to handle. & Correct shortest-path action \\
\midrule
State 4 & A & Power down the television to eliminate any screen glare during the movie. & Invalid in current state \\*
State 4 & B & Walk up to the TV unit --- the cabinet is right ahead. & Correct shortest-path action \\*
State 4 & C & Focus on the television to assess whether the screen needs attention. & Invalid in current state \\*
State 4 & D & Switch on the television so you can test the viewing angle for the movie. & Invalid in current state \\
\midrule
State 5 & A & Shift your focus to the ceiling light to confirm it's off for movie time. & Executable detour \\*
State 5 & B & Open the media cabinet so everything is within reach for the movie. & Correct shortest-path action \\*
State 5 & C & Take a closer look at the television area to plan the next move for the media setup. & Invalid in current state \\*
State 5 & D & Step over to the television unit to get a better position for the entertainment area. & Executable detour \\
\end{longtable}
\endgroup

\begingroup
\setlength{\tabcolsep}{3pt}
\renewcommand{\arraystretch}{1.15}
\setlength{\LTcapwidth}{\textwidth}
\setlength{\LTpre}{0pt}
\setlength{\LTpost}{1em}
\begin{longtable}{@{}>{\scriptsize\raggedright\arraybackslash}p{\dimexpr0.095\textwidth-3pt\relax}>{\scriptsize\raggedright\arraybackslash}p{\dimexpr0.08\textwidth-3pt\relax}>{\scriptsize\raggedright\arraybackslash}p{\dimexpr0.585\textwidth-6pt\relax}>{\scriptsize\raggedright\arraybackslash}p{\dimexpr0.24\textwidth-6pt\relax}@{}}
\caption{Basic-task example: action options and hidden graph roles at each state.}
\vspace{-2mm}
\label{tab:tongbench-l2-example-0022-options} \\
\toprule
\multicolumn{4}{@{}p{\textwidth}@{}}{\scriptsize \textbf{Task:} Get the living room ready for a relaxed evening --- turn on some background entertainment and open up the storage cabinet so everything's within reach.} \\
\midrule
\textbf{State} & \textbf{Option} & \textbf{Text shown to the agent} & \textbf{Hidden role} \\
\midrule
\endfirsthead
\multicolumn{4}{@{}l@{}}{\normalsize Table~\thetable\ (continued).} \\[0.5em]
\toprule
\textbf{State} & \textbf{Option} & \textbf{Text shown to the agent} & \textbf{Hidden role} \\
\midrule
\endhead
\midrule
\multicolumn{4}{@{}r@{}}{\scriptsize\textit{Continued on the next page.}} \\
\endfoot
\bottomrule
\endlastfoot
State 0 & A & Start by checking the TV area --- that's where the background entertainment will come from. & Correct shortest-path action \\*
State 0 & B & Shut off the TV screen since it's not needed for the living room ambiance. & Invalid in current state \\*
State 0 & C & Head over to the TV wall to get closer to the entertainment controls. & Invalid in current state \\*
State 0 & D & Pull open the storage cabinet to make everything accessible for the evening. & Invalid in current state \\
\midrule
State 1 & A & Shut the storage cabinet to keep the living room looking tidy. & Invalid in current state \\*
State 1 & B & Shift your focus to the cabinet area near the TV to assess what's needed. & Invalid in current state \\*
State 1 & C & Get to the entertainment setup --- the background noise needs to start soon. & Correct shortest-path action \\*
State 1 & D & Pull open the storage compartment so things can be reached easily for the evening. & Invalid in current state \\
\midrule
State 2 & A & Switch on the screen to bring some background entertainment into the living room. & Correct shortest-path action \\*
State 2 & B & Head over to the screen area to get a better position for the setup. & Executable detour \\*
State 2 & C & Focus on the storage cabinet to make sure it's ready for easy access. & Executable detour \\*
State 2 & D & Check the screen to see if it's already running before proceeding. & Invalid in current state \\
\midrule
State 3 & A & The storage cabinet still needs opening --- that's the next thing to handle. & Correct shortest-path action \\*
State 3 & B & Head over to the storage unit to get it ready for the evening. & Invalid in current state \\*
State 3 & C & Switch off the ambient background element near the TV to set a cleaner mood. & Executable detour \\*
State 3 & D & Walk toward the entertainment area to adjust the atmosphere. & Executable detour \\
\midrule
State 4 & A & Shut the storage cabinet to keep the living room looking tidy. & Invalid in current state \\*
State 4 & B & Check the cabinet area to see what's currently inside. & Invalid in current state \\*
State 4 & C & Pull open the storage cabinet to make everything accessible for the evening. & Invalid in current state \\*
State 4 & D & Move into position near the storage cabinet --- it needs to be opened. & Correct shortest-path action \\
\midrule
State 5 & A & Step closer to the storage cabinet to get a better grip on the handle. & Executable detour \\*
State 5 & B & Scan the cabinet area to confirm everything is in order. & Invalid in current state \\*
State 5 & C & Glance toward the TV wall to take in the overall setup before finishing up. & Executable detour \\*
State 5 & D & Open up the storage cabinet so everything is within reach for the evening. & Correct shortest-path action \\
\end{longtable}
\endgroup

\begingroup
\setlength{\tabcolsep}{3pt}
\renewcommand{\arraystretch}{1.15}
\setlength{\LTcapwidth}{\textwidth}
\setlength{\LTpre}{0pt}
\setlength{\LTpost}{1em}
\begin{longtable}{@{}>{\scriptsize\raggedright\arraybackslash}p{\dimexpr0.095\textwidth-3pt\relax}>{\scriptsize\raggedright\arraybackslash}p{\dimexpr0.08\textwidth-3pt\relax}>{\scriptsize\raggedright\arraybackslash}p{\dimexpr0.585\textwidth-6pt\relax}>{\scriptsize\raggedright\arraybackslash}p{\dimexpr0.24\textwidth-6pt\relax}@{}}
\caption{Advanced-task example: action options and hidden graph roles.}
\vspace{-2mm}
\label{tab:tongbench-l4-example-options} \\
\toprule
\multicolumn{4}{@{}p{\textwidth}@{}}{\scriptsize \textbf{Task:} I'm done for the night and heading to bed --- dim the bedroom lamp, open up the study desk so it's ready for tomorrow, shut the bedside drawer, and toss the old remote into the bin.} \\
\midrule
\textbf{State} & \textbf{Option} & \textbf{Text shown to the agent} & \textbf{Hidden role} \\
\midrule
\endfirsthead
\multicolumn{4}{@{}l@{}}{\normalsize Table~\thetable\ (continued).} \\[0.5em]
\toprule
\textbf{State} & \textbf{Option} & \textbf{Text shown to the agent} & \textbf{Hidden role} \\
\midrule
\endhead
\midrule
\multicolumn{4}{@{}r@{}}{\scriptsize\textit{Continued on the next page.}} \\
\endfoot
\bottomrule
\endlastfoot
State 0 & A & Check the trash can area to see what needs to be tossed before bed. & Executable detour \\*
State 0 & B & Switch on the table lamp to brighten the bedside area for cleanup. & Invalid in current state \\*
State 0 & C & The lamp's still glowing --- that's the first thing to sort out before the room can rest. & Correct shortest-path action \\*
State 0 & D & Head over to the table lamp to adjust the lighting for bedtime. & Invalid in current state \\
\midrule
State 1 & A & Switch off the table lamp to dim the room for bedtime. & Invalid in current state \\*
State 1 & B & Get closer to the lamp --- you'll need to be right there to kill the light. & Correct shortest-path action \\*
State 1 & C & Brighten the table lamp so you can see what's left to tidy. & Invalid in current state \\*
State 1 & D & Focus on the table lamp area to assess what needs adjusting for the night. & Invalid in current state \\
\midrule
State 2 & A & Brighten the lamp a bit so you can see what's left on the nightstand before closing it up. & Invalid in current state \\*
State 2 & B & Kill the lamp light to start dimming the room for the night. & Correct shortest-path action \\*
State 2 & C & Shift over to the lamp side to get the lighting sorted before settling in. & Invalid in current state \\*
State 2 & D & Focus on the lamp area to assess what still needs adjusting before bed. & Invalid in current state \\
\midrule
State 3 & A & Glance over at the trash can to see what needs tossing before bed. & Executable detour \\*
State 3 & B & The study desk still has stuff on it --- that's the next area to sort out. & Correct shortest-path action \\*
State 3 & C & Head over to the study desk to get in position for clearing it off. & Invalid in current state \\*
State 3 & D & Flip open the study desk to start organizing items for tomorrow's setup. & Invalid in current state \\
\midrule
State 4 & A & Glance over at the desk to see what's still out on the surface. & Invalid in current state \\*
State 4 & B & Flip open the desk to start organizing what's inside for tomorrow. & Invalid in current state \\*
State 4 & C & Shut the desk drawers to keep the room tidy for the night. & Invalid in current state \\*
State 4 & D & Step up to the desk so you can reach across and clear it off properly. & Correct shortest-path action \\
\midrule
State 5 & A & Shift your focus to the desk to assess what still needs clearing. & Invalid in current state \\*
State 5 & B & Open the desk cabinet so you can stow tonight's items inside. & Correct shortest-path action \\*
State 5 & C & Glance over at the table lamp to check whether the room is dim enough. & Executable detour \\*
State 5 & D & Take a moment to review the bedside cabinet before closing it up for the night. & Executable detour \\
\midrule
State 6 & A & Shift your focus to the table lamp to see if it still needs attention for the night. & Executable detour \\*
State 6 & B & Glance over at the trash can to check if anything else needs tossing before bed. & Executable detour \\*
State 6 & C & Take a look at the remote control in your hand to decide where it should go next. & Invalid in current state \\*
State 6 & D & The bedroom nightstand is still open --- head back and deal with that next. & Correct shortest-path action \\
\midrule
State 7 & A & Shut the bedside cabinet to keep the room tidy for the night. & Invalid in current state \\*
State 7 & B & Pull the bedside cabinet open to check if anything needs to be put away. & Invalid in current state \\*
State 7 & C & Glance over at the bedside cabinet to see what's still out. & Invalid in current state \\*
State 7 & D & Move in close to the nightstand --- the cabinet needs shutting and you need to be right there. & Correct shortest-path action \\
\midrule
State 8 & A & Shift focus to the desk area to see what needs clearing for tomorrow. & Executable detour \\*
State 8 & B & Shut the bedside cabinet to keep the nightstand tidy before sleep. & Correct shortest-path action \\*
State 8 & C & Check the table lamp status to see if it needs dimming for bed. & Executable detour \\*
State 8 & D & Glance at the bedside cabinet to confirm what's inside before closing. & Invalid in current state \\
\midrule
State 9 & A & Shift focus to the table lamp area to continue dimming the room for rest. & Executable detour \\*
State 9 & B & The remote control still needs to go --- spot the trash can so you know where it's headed. & Correct shortest-path action \\*
State 9 & C & Glance over at the desk surface to assess what else still needs to be cleared away. & Executable detour \\*
State 9 & D & Check the remote control in your hand to decide whether it should be put away. & Invalid in current state \\
\midrule
State 10 & A & Seal up the trash can so nothing spills out overnight. & Invalid in current state \\*
State 10 & B & Glance at the trash can to make sure it's ready to receive items. & Invalid in current state \\*
State 10 & C & Pop open the trash can lid so you can drop things in. & Invalid in current state \\*
State 10 & D & Get to the trash can --- the remote control in your hand needs to go there. & Correct shortest-path action \\
\midrule
State 11 & A & Set the remote control down near the trash can to keep the nightstand clear. & Correct shortest-path action \\*
State 11 & B & Check the bedside cabinet to make sure it's properly closed for the night. & Executable detour \\*
State 11 & C & Take a closer look at the remote control to decide where it should go. & Invalid in current state \\*
State 11 & D & Turn your attention to the desk area to see what still needs clearing. & Executable detour \\
\end{longtable}
\endgroup

\end{document}